\documentclass[]{imsart}

\RequirePackage[OT1]{fontenc}
\RequirePackage{amsthm,amsmath,amssymb,bbm}
\RequirePackage[numbers]{natbib}
\RequirePackage[colorlinks,citecolor=blue,urlcolor=blue]{hyperref}
\usepackage{subcaption}
\usepackage[ruled]{algorithm2e}

\startlocaldefs
\numberwithin{equation}{section}
\theoremstyle{plain}

\endlocaldefs

\makeatletter
\def\singlespace{\def\baselinestretch{1}\@normalsize}

\makeatletter
\def\singlespace{\def\baselinestretch{1}\@normalsize}

\numberwithin{equation}{section}

\renewcommand{\hat}{\widehat}

\renewcommand{\hat}{\widehat}

     \def\CC{\mathbb{C}}
     \def\DD{\mathbb{D}}
     \def\EE{\mathbb{E}}

     \def\HH{\mathbb{H}}

     \def\NN{\mathbb{N}}
     
     \def\PP{\mathbb{P}}
     
     \def\RR{\mathbb{R}}

\newcommand{\bfsym}[1]{\ensuremath{\boldsymbol{#1}}}

           \def\bTheta {\bfsym {\Theta}}

\allowdisplaybreaks
\usepackage{verbatim}
\newtheorem{dfn}{Definition}
\newcounter{CondCounter}
\numberwithin{equation}{section}
\theoremstyle{plain}
\newtheorem{theorem}{Theorem}[section]

\theoremstyle{definition}
\newtheorem{remark}{Remark}	

\newtheorem{proposition}[theorem]{Proposition}

\theoremstyle{definition}

\theoremstyle{definition}
\newtheorem{lemma}{Lemma}

\newtheorem{corollary}{Corollary}[section]

\theoremstyle{definition}

\theoremstyle{definition}

\theoremstyle{definition}

\usepackage{mathtools}
\DeclarePairedDelimiter\ceil{\lceil}{\rceil}

\begin{document}

\begin{frontmatter}
\title{Nonparametric Estimation of Low Rank Matrix Valued Function}
\runtitle{Nonparametric Estimation of Low Rank Matrix Valued Function}

\begin{aug}
\author{\fnms{Fan} \snm{Zhou}\thanksref{t2}\ead[label=e1]{fzhou40@math.gatech.edu}}

\address{School of Mathematics\\
Georgia Institute of Technology\\
Atlanta, GA 30332-0160\\
\printead{e1}}

\thankstext{t2}{Supported in part by NSF Grants DMS-1509739 and CCF-1523768.}
\runauthor{Fan Zhou}

\affiliation{Georgia Institute of Technology}
\tableofcontents
\end{aug}

\begin{abstract}
Let $A:[0,1]\rightarrow\mathbb{H}_m$ (the space of Hermitian matrices) be a matrix valued function which is low rank with entries in H\"{o}lder class $\Sigma(\beta,L)$.
The goal of this paper is to study statistical estimation of $A$ based on the regression model
$
\EE(Y_j|\tau_j,X_j) = \langle A(\tau_j), X_j \rangle,
$
where $\tau_j$ are i.i.d. uniformly distributed in $[0,1]$, $X_j$ are i.i.d. matrix completion sampling matrices, $Y_j$ are independent bounded responses. 
We propose an innovative nuclear norm penalized local polynomial estimator and establish an upper bound on its point-wise risk measured by Frobenius norm. Then we extend this 
estimator globally and prove an upper bound on its integrated risk measured by $L_2$-norm. We also propose another new estimator based on bias-reducing kernels to study the case when 
$A$ is not necessarily low rank and establish an upper bound on its risk measured by $L_{\infty}$-norm. We show that the obtained rates are all optimal up to some logarithmic factor in minimax sense. Finally, we propose an adaptive estimation procedure based on Lepski's method and the penalized data splitting technique which is computationally efficient and can be easily implemented and
parallelized on distributed systems.
\end{abstract}


\end{frontmatter}

\section{Introduction}
Let $A:[0,1]\rightarrow\mathbb{H}_m$ (the space of Hermitian matrices)\footnote{Note that we use $\HH_m$ for the simplicity of presentation, and our results can be trivially generalize to regular matrix spaces such as $\RR^{m_1\times m_2}$.} be a matrix valued function. The goal of this paper is to study the problem of statistical estimation of a matrix valued function $A$ based on the regression model
\begin{equation}
\EE(Y_j | \tau_j,X_j) = \langle A(\tau_j), X_j \rangle,~~ j=1,\ldots,n, \label{Model}
\end{equation}
where $\tau_j$ are i.i.d. random univariates uniformly distributed on $[0,1]$, $X_j$ are i.i.d. matrix completion sampling matrices, $Y_j$ are independent bounded random responses. 
Sometimes, it is convenient to write model (\ref{Model}) in the form
\begin{equation}
Y_j = \langle A(\tau_j), X_j \rangle + \xi_j,~~ j=1,\ldots,n, \label{Model1}
\end{equation}
where the noise variables $\xi_j = Y_j - \EE(Y_j | \tau_j,X_j)$ are independent and have zero means.
In particular, we are interested in the case where $A$ is low rank and its entries belong to a standard function class $\Sigma(\beta, L)$ which is called H\"{o}lder class, see Definition \ref{def:def_2}. When $A(t) = A_0$ with some fixed $A_0\in \HH_m$ for any $t \in [0,1]$, such a problem coincides with the well known matrix completion/recovery problem that
has drawn a lot of attention in the statistics community during the past few years, see \cite{candes2009exact,cai2010singular,candes2010matrix,candes2010power,koltchinskii2011neumann,gross2011recovering,koltchinskii2011nuclear,rohde2011estimation,negahban2012restricted,chatterjee2015matrix} and the references therein.
The low rank assumption in matrix completion/estimation problems has profound practical background. In the following, we discuss several simple examples of general low rank matrix valued functions that fit in our problem.

Example 1. Dynamic~Collaborative~Filtering~Model.  Let $P: [0,1]\rightarrow \RR^{m_1\times r}$ and $Q:[0,1]\rightarrow \RR^{m_2\times r}$, then $A = PQ^T$ is apparently a low rank matrix valued function $r \ll (m_1 \wedge m_2)$. This kind of dynamic collaborative filtering model was initially introduced by \cite{koren2010collaborative}, which generalized their well known work \cite{koren2009matrix} to tackle the Netflix Prize on building 
dynamic recommender systems.

Example 2. Matrix~Function~Multiplication~Model. Let $f:[0,1] \rightarrow \RR$ and $\tilde{A} \in \RR^{m_1\times m_2}$, where ${\rm rank}(\tilde{A}) = r\ll (m_1 \wedge m_2)$. Then $A = \tilde{A}*f$ is a low rank matrix valued function. Such applications can be found in biology, chemistry and signal processing (see \cite{weickert2002diffusion,miller1976classical}) where the underlying information $\tilde{A}$ is diffused via certain uniform diffusion function $f$.

Example 3. Euclidean~Distance~Matrix~Model. Given the trajectory vectors of $m$ points in $\RR^d$, $p_1,...,p_m : [0,1]\rightarrow \RR^d$. Then the Euclidean distance matrix (EDM) $D:[0,1] \rightarrow \RR^{m\times m}$
with $D_{ij} = \|p_i - p_j\|_2^2$ is a matrix valued function with rank at most $d+2$ regardless of its size $m$. Clearly, when $m \gg d$, $D$ falls into the low rank realm. In molecular biology, such points are typically 
in a low dimensional space such as $\RR^2$ or $\RR^3$. Similar topics in cases when points are fixed (see \cite{zhang2016distance}) or in rigid motion (see \cite{singer2010uniqueness}) have been studied.

An appealing way to address the low rank issue in matrix recovery problems is through nuclear norm minimization, see \cite{recht2010guaranteed}. In section 3, we inherit this idea and propose a local polynomial estimator (see \cite{fan1996local}) with nuclear norm penalization:
\begin{equation}
\label{point_estimator_1}
\hat{S}^h =\arg\min\limits_{S\in \DD} \frac{1}{nh}\sum\limits_{j=1}^{n}K\Big(\frac{\tau_j-t_0}{h}\Big)\Big(Y_j-\Big\langle \sum\limits_{i=0}^{\ell}S_i p_i\Big(\frac{\tau_j-t_0}{h}\Big),X_j\Big\rangle\Big)^2 +\varepsilon \|S\|_1. 
\end{equation}
where $\DD \subset \HH_{(\ell+1)m}$ is a closed subset of block diagonal matrices with $S_j\in \HH_{m}$ on its diagonal, and $\{p_i(t)\}_{i=0}^{\infty}$ is a sequence of orthogonal polynomials with nonnegative weight 
function $K$. The solution to the convex optimization problem (\ref{point_estimator_1}) induces a pointwise estimator of $A(t_0)$: 
$$
\hat{S}^{h}(t_0):=\sum_{i=0}^{\ell}\hat{S}^h_i p_i(0)
$$
where $\hat{S}^h_i$ are the blocks on the diagonal of $\hat{S}^h$ and $\ell = \lfloor \beta\rfloor$.
We prove that under mild conditions, the pointwise risk measured by $m^{-2}\big\|\hat{S}^{h}(t_0)-A(t_0)\big\|_2^2$ of $\hat{S}^{h}(t_0)$ over H\"{o}lder class $\Sigma(\beta,L)$ satisfies the following upper bound
\begin{equation}
\label{rate: fro}
	m^{-2}\big\|\hat{S}^{h}(t_0)-A(t_0)\big\|_2^2 = O_p\Big(\Big[\frac{mr\log n}{n}\Big]^\frac{2\beta}{2\beta+1}\Big)
\end{equation}
where $r$ is the low rank parameter and $\|\cdot\|_2$ denotes the Frobenius norm of a matrix.

In section 4, we propose a new global estimator $\hat{A}$ based on local polynomial smoothing and prove that the integrated risk of $\hat{A}$ measured by $L_2$-norm satisfies the following upper bound
\begin{equation}
\label{rate: l2}
	m^{-2}\int_0^1 \big\| \hat{A}(t) - A(t)\big\|_2^2dt = O_p\Big(\Big[\frac{mr\log n}{n}\Big]^{\frac{2\beta}{2\beta+1}}\Big).
\end{equation}
Then we study another naive kernel estimator $\tilde{A}$ which can be used to estimate matrix valued functions which are not necessarily low rank. This estimator is associated with another popular approach to deal with low rank recovery which is called singular value thresholding, see \cite{cai2010singular,koltchinskii2011nuclear,chatterjee2015matrix}. 
We prove that the $\sup$-norm risk of $\tilde{A}$ satisfies the following upper bound
\begin{equation}
\label{rate: sup}
	\sup_{t\in[h,1-h]} m^{-2} \big\|\tilde{A}(t)-A(t)\big\|^2 = O_p\Big(\Big[\frac{m\log n}{n}\Big]^{\frac{2\beta}{2\beta+1}}\Big),
\end{equation}
where $\|\cdot\|$ denotes the matrix operator norm. Note that those rates coincide with that of classical matrix recovery setting when the smoothness parameter $\beta$ goes to infinity.

An immediate question to ask is whether the above rates are optimal. In section 5, we prove that 
the rates in (\ref{rate: fro}), (\ref{rate: l2}) and (\ref{rate: sup}) are all optimal up to some logarithmic factor in minimax sense, which essentially verified the effectiveness of our methodology.

As one may have noticed, there is an adaptation issue involved in (\ref{point_estimator_1}). Namely, one needs to choose a proper bandwidth $h$ and a proper order of degree $\ell$ of polynomials. 
Both parameters are closely related to the smoothness parameter $\beta$ of $A$ which is unknown to us in advance. In section 6, we propose a model selection procedure based on Lepskii's method (\cite{lepskii1991problem}) and the work of \cite{barron1991complexity} and \cite{wegkamp2003model}. We prove that this procedure adaptively selects an estimator $\hat{A}^*$ such that the integrated risk of 
$\hat{A}^*$ measured by $L_2$-norm has the following upper bound 
\begin{equation}
		m^{-2}\int_0^1 \big\| \hat{A}^*(t) - A(t)\big\|_2^2dt = O_p\Big(\Big[\frac{mr\log n}{n}\Big]^{\frac{2\beta}{2\beta+1}}\Big)
\end{equation}
which is still near optimal. What is more important, such a procedure is computationally efficient, feasible in high dimensional setting, and can be easily parallelized.

The major contribution of this paper is on the theory front. We generalized the recent developments of low rank matrix completion theory to nonparametric estimation setting by proposing an innovative optimal estimation procedure. To our best knowledge, no one has ever thoroughly studied such problems from a theoretical point of view. 

\section{Preliminaries}
\label{preliminary}
In this section, we introduce some important definitions, basic facts, and notations for the convenience of presentation.

\subsection{Notations}
For any Hermitian matrices $A, B\in \mathbb{H}_m$, denote $\langle A, B \rangle = {\rm tr}(AB)$
which is known as the Hilbert-Schmidt inner product. Denote $\langle A, B \rangle_{L_2(\Pi)} =\mathbb{E}\langle A, X\rangle \langle B, X\rangle$,
where $\Pi$ denotes the distribution of $X$. The corresponding norm $\| A \|^2_{L_2(\Pi)}$ is given by $\| A \|_{L_2(\Pi)}^2=\mathbb{E}\langle A, X\rangle^2.$

We use $\| \cdot \|_2$ to denote the Hilbert-Schimidt norm (Frobenus norm or Schatten 2-norm) induced by the inner product $\langle \cdot, \cdot \rangle$;
$\| \cdot \|$ to denote the operator norm (spectral norm) of a matrix: the largest singular value; $\| \cdot \|_1$ to denote the trace norm (Schatten 1-norm or nuclear norm), i.e. the sum of singular values; $|A|$ to denote the nonnegative matrix with entries $|A_{ij}|$ corresponding to $A$.

Given $X_1$,...,$X_n$ as the i.i.d. copies of the random measurement matrix $X$, denote
$$
\sigma^2_{X}:=\Big\|n^{-1} \sum\limits_{j=1}^n \mathbb{E}X^2_{i}\Big\|,~U_{X}:= \big{\|}\|X\|\big{\|}_{L_\infty}.
$$
where $U_X$ denotes the $L_{\infty}$-norm of the random variable $\|X\|$.
\subsection{Matrix completion and statistical learning setting}
The matrix completion setting refers to that the random sampling matrices $X_j$ are i.i.d. uniformly distributed on the following orthonormal basis $\mathcal{X}$ of $\HH_m$:
$$
\mathcal{X} :=\{E_{kj}:k,j=1,...,m\},
$$
where $E_{kk}:=e_k\otimes e_k$, $k=1,...,m$; $E_{jk}:=\frac{1}{\sqrt{2}}(e_k\otimes e_j+e_j\otimes e_k)$, $1\leq k<j\leq m$; $E_{kj}:=\frac{i}{\sqrt{2}}(e_k\otimes e_j-e_j\otimes e_k)$, 
$1\leq k<j\leq m$ with $\{e_j\}_{j=1}^m$ being the canonical basis of $\RR^m$.
The following identities are easy to check when the design matrices are under matrix completion setting:
\begin{equation}
\label{identity}
\big\|A\big\|^2_{L_2(\Pi)}=\frac{1}{m^2}\big\|A\big\|_2^2,~~~\sigma^2_X \leq \frac{2}{m},~~~U_X =1.
\end{equation}
The statistical learning setting refers to the bounded response case: there exists a constant $a$ such that 
\begin{equation}
\max_{j=1,...n}|Y_j|\leq a,~a.s. \label{statistical learning}
\end{equation}
In this paper, we will consider model (\ref{Model}) under both matrix completion and statistical learning setting.

\subsection{Matrix valued function}
Let	$A: [0,1]\rightarrow \mathbb{H}_m$ be a matrix valued function. One should notice that we consider the image space to be Hermitian matrix space for the convenience of presentation. Our methods and results can be readily extended to general rectangular matrix space. Now we define the rank of a matrix valued function.
Let 
$
{\rm rank}_A(t) := {\rm rank}(A(t)),~\forall t\in [0,1].
$

\begin{dfn}
	\label{def:def_2}
	Let $\beta$ and $L$ be two positive real numbers. The \textbf{H\"{o}lder class}
	$\Sigma(\beta,L)$ on $[0,1]$ is defined as the set of $\ell=\lfloor \beta\rfloor$ times differentiable functions $f:[0,1] \rightarrow \mathbb{R}$ with derivative 
	$f^{(\ell)}$ satisfying
	\begin{equation}
	|f^{(\ell)}(x)-f^{(\ell)}(x')| \leq L|x-x'|^{\beta-\ell},~~\forall x,x'\in [0,1].
	\end{equation}
\end{dfn}
The parameters $\beta$ and $\ell$ characterize the smoothness of H\"{o}lder class $\Sigma(\beta, L)$. They are the most important parameters in our problem just like the dimension of the matrix $m$ and sample
size $n$. Throughout this paper, we only consider the case when $\ell$ is a fixed constant, or in other words $\ell \ll m$. The reason is that in the asymptotic theory of low rank matrix recovery, the size of $m$ is often considered to be comparable to the sample size $n$, say $m=O(n)$. If $\ell$ is also comparable to $m$, then the our theory can be problematic.

In particular, we are interested in the following assumptions on matrix valued functions:
\begin{description}
	\item[A1] Given a measurement matrix $X$ and for some constant $a_1$,
	$$
	\sup\limits_{t\in [0,1]}\big|\langle A(t),X\rangle\big| \leq a_1.
	$$
	
	\item[A2] Given a measurement matrix $X$ and for some constant $a_2$, the derivative matrices $A^{(k)}$ of $A$ satisfy
	$$
	\sup\limits_{t\in [0,1]}\big|\langle A^{(k)}(t),X\rangle\big| \leq a_2,~~k=1,...,\ell.
	$$
	\item[A3] The rank of $A$, $A'$, ...,$A^{(\ell)}$ are uniformly bounded by a constant $r$, 
	$$
	\sup\limits_{t\in [0,1]}{\rm rank}_{A^{(k)}}(t)\leq r,~~k=0,1,...,\ell.
	$$
	\item[A4] Assume that for $\forall i,j$, the entry $A_{ij}$ is in the H\"{o}lder class $\Sigma(\beta,L)$.
\end{description}

\section{A Local Polynomial Lasso Estimator}
\label{section: local_lasso}
In this section, we study the pointwise estimation of a low rank matrix valued function $A$ in $\Sigma(\beta, L)$
with $\ell = \lfloor \beta \rfloor$.
The construction of our estimator is inspired by local polynomial smoothing and nuclear norm penalization.
The intuition of the localization technique originates from classical local polynomial estimators, see \cite{fan1996local}. 
The intuition behind nuclear norm penalization is that whereas rank function counts the number of non-vanishing singular values, the nuclear norm sums their amplitude.
The theoretical foundations behind nuclear norm heuristic for the rank minimization were proved by \cite{recht2010guaranteed}. 
Instead of using the trivial basis $\{1,t,t^2,...,t^{\ell}\}$ to generate an estimator, we use orthogonal polynomials. Let $\{p_i(t)\}_{i=0}^\infty$ be a sequence of orthogonal polynomials with nonnegative weight function $K$ compactly supported on $[-1,1]$, then
$$
\int_{-1}^1 K(u) p_i(u) p_j (u) du = \delta_{ij}
$$
with $\delta_{ij} = \mathbf{1}\{i=j\}$.
There exists an invertible linear transformation $T\in \RR^{(\ell+1)\times (\ell+1)}$ such that 
$$
(1,t,t^2/2!,...,t^{\ell}/\ell!)^T = T(p_0,p_1,...,p_{\ell})^T.
$$
Apparently, $T$ is lower triangular. We denote $R(T)=\max_{1\leq j \leq \ell+1}\sum_{i=1}^{\ell+1}|T_{ij}|$. 

Denote 
$$
\DD
:= \Big\{{\rm Diag}
\begin{bmatrix}
S_0 & S_1 & \dots & S_{\ell-1} & S_{\ell} \\
\end{bmatrix} \Big\} \subset \HH_{m(\ell+1)}
$$
the set of block diagonal matrices with $S_k \in \HH_m$ satisfying $|S_{ij}| \leq R(T)a$.
With observations $(\tau_j,X_j,Y_j)$, $j=1,...,n$ from model (\ref{Model}), define $\hat{S}^h$ as
\begin{equation}
\label{Estimator}
\hat{S}^h :=\arg\min\limits_{S\in \DD} \frac{1}{nh}\sum\limits_{j=1}^{n}K\Big(\frac{\tau_j-t_0}{h}\Big)\Big(Y_j-\Big\langle \sum\limits_{i=0}^{\ell}S_i p_i\Big(\frac{\tau_j-t_0}{h}\Big),X_j\Big\rangle\Big)^2 +\varepsilon \|S\|_1. 
\end{equation}
\begin{remark}
	Note that one can rewrite (\ref{Estimator}) as 
	\begin{equation}
	\label{Reform:Estimator}
	\hat{S}^h =\arg\min\limits_{S\in \mathbb{D}} \frac{1}{n}\sum\limits_{j=1}^{n} \Big(\tilde{Y}_j-\Big\langle  S , \tilde{X}_j\Big\rangle\Big)^2 +\varepsilon \|S\|_1. 
	\end{equation}
	where $\tilde{X}_j = {\rm Diag}\Big[\sqrt{\frac{1}{h}K\Big(\frac{\tau_j-t_0}{h}\Big)}p_0\Big(\frac{\tau_j-t_0}{h}\Big) X_j,...,\sqrt{\frac{1}{h}K\Big(\frac{\tau_j-t_0}{h}\Big)}p_{\ell}\Big(\frac{\tau_j-t_0}{h}\Big)X_j\Big]$, and
	$\tilde{Y}_j = \sqrt{\frac{1}{h}K\Big(\frac{\tau_j-t_0}{h}\Big)}Y_j$. Then (\ref{Estimator}) is a matrix Lasso type estimator. 
\end{remark}

$\hat{S}^h$ naturally induces a local polynomial estimator of order $\ell$ around $t_0$: 
\begin{equation}
\label{local_poly}
\hat{S}^h(\tau): = \sum\limits_{i=0}^{\ell} \hat{S}^h_i p_i\Big(\frac{\tau-t_0}{h}\Big)\mathbf{1}\Big\{\Big|\frac{\tau-t_0}{h}\Big|\leq 1\Big\}.
\end{equation}
The point estimate of $A$ at $t_0$ is given by
\begin{equation}
\hat{S}^h(t_0) := \sum\limits_{i=0}^{\ell} \hat{S}^h_i p_i(0). \label{point_estimator}
\end{equation}
\begin{remark}
	Note that (\ref{Estimator}) only guarantees that each $\hat{S}^h_i$ is approximately low rank and may not exactly recover the rank of $A^{(i)}(t_0)$. However, under our assumption that as long as $\ell$ is small compared with the matrix size $m$, then $\hat{S}^h(t_0)$ is still approximately low rank.
\end{remark}

In the following theorem, we establish an upper bound on the pointwise risk of $\hat{S}^h(t_0)$ when $A(t)$ is in the H\"{o}lder class $\Sigma(\beta,L)$ with $\ell = \lfloor \beta\rfloor$. The proof of Theorem \ref{Thm_2} can be found in section \ref{proof:Thm_2}. 

\begin{theorem}
	\label{Thm_2}
	Under model (\ref{Model}),
	let $(\tau_j, X_j, Y_j)$, $j=1,...,n$ be i.i.d. copies of the random triplet $(\tau,X,Y)$ with
	$X$ uniformly distributed in $\mathcal{X}$, $\tau$ uniformly distributed in $[0,1]$, $X$ and $\tau$ are independent, and 
	$|Y|\leq a$, a.s. for some constant $a>0$.
	Let $A$ be a matrix valued function satisfying A1, A2, A3, and A4. Denote $\Phi =\max_{i=0,...,\ell} \|\sqrt{K} p_i\|_{\infty}$, and $\ell=\lfloor\beta\rfloor$.
	Take 
	$$
	\hat{h}_n = C_1 \Big( \frac{(\ell^3 (\ell !)^2\Phi^2 R(T)^2 a^2 mr\log n }{L^2n} \Big)^{\frac{1}{2\beta+1}},~\varepsilon = D\ell a\Phi\sqrt{\frac{\log 2m}{nm\hat{h}_n }},
	$$
	for some numerical constants $C_1$ and $D$.
	Then for any $\hat{h}_n \leq t_0 \leq 1-\hat{h}_n$, the following bound holds with probability at least $1-n^{-mr}$,
	\begin{equation} 
	\label{Thm_bound_3}
	\frac{1}{m^2} \Big\| \hat{S}^h(t_0)- A(t_0)\Big\|_2^2 \leq C_1(a,\Phi,\ell,L) \Big(\frac{mr\log n}{n}\Big)^{\frac{2\beta}{2\beta+1}}, 
	\end{equation}
	where $C_1(a,\Phi,\ell,L)$ is a constant depending on $a,\Phi,\ell$ and $L$.
\end{theorem}
\begin{remark}
	One should notice that when $\beta \rightarrow \infty$, bound (\ref{Thm_bound_3}) coincides with the similar result 
	in classical matrix completion of which the rate is $O_p\big( \frac{mr\log m}{n}\big)$, see \cite{koltchinskii2011nuclear}. As long as $n$ is of the polynomial order of $m$, there is only up to a constant between
	$\log n$ and $\log m$. In section \ref{Lower_bounds}, we prove that bound (\ref{Thm_bound_3}) is minimax optimal up to a logarithmic factor. The logarithmic factor in bound (\ref{Thm_bound_3}) and bound of classical matrix completion is introduced by matrix Bernstein inequality, see \cite{tropp2012user}. In the case of nonparametric estimation of real valued function, it is unnecessary, see \cite{tsybakov2009introduction}.
\end{remark}

\section{Global Estimators and Upper Bounds on Integrated Risk}
\label{section: global}
In this section, we propose two global estimators and study their integrated risk measured by $L_2$-norm and $L_{\infty}$-norm. 
\subsection{From localization to globalization}
Firstly, we construct a global estimator based on (\ref{local_poly}).
Take 
$$
\hat{h}_n = C_1 \Big( \frac{\ell^3 (\ell !)^2\Phi^2 R(T)^2 a^2 mr\log n }{L^2n} \Big)^{\frac{1}{2\beta+1}},~M= \lceil1/\hat{h}_n \rceil.
$$
Without loss of generality, assume that $M$ is even.
Denote $\hat{S}_k^h(t)$ the local polynomial estimator around $t_{2k-1}$ as in (\ref{local_poly}) by using orthogonal polynomials with $K(t)=\mathbf{1}\{-1\leq t \leq 1\}$, where $t_{2k-1} = \frac{2k-1}{M}$, $k=1,2,...,M/2$ and 
$\mathbf{1}\{\cdot  \}$ is the indicator function. 
Denote
\begin{equation}
\hat{A}(t) = \sum\limits_{k=1}^{M/2} \hat{S}_k^h(t)\mathbf{1}\{ t_{2k-1}-\hat{h}_n< t \leq t_{2k-1}+\hat{h}_n \},~~t\in(0,1). \label{Global_estimator}
\end{equation}
Note that the weight function $K$ is not necessary to be $\mathbf{1}\{-1\leq t \leq 1\}$. It can be replaced by any $K$ that satisfies $K\geq K_0>0$ on $[-1,1]$.
The following result characterizes the integrated risk of estimator (\ref{Global_estimator}) under matrix completion setting measured by $L_2$-norm. The proof of Theorem \ref{Thm_4} can be found in section \ref{ProofThm_4}.

\begin{theorem}
	\label{Thm_4}
	Assume that the conditions of Theorem \ref{Thm_2} hold, and let $\hat{A}$ be an estimator defined as in (\ref{Global_estimator}).
	Then with probability at least $1-n^{-(mr-1)}$,
	\begin{equation}
	\frac{1}{m^2}\int^1_0 \big\|\hat{A}(t)-A(t)\big\|_2^2 dt \leq C_2(a,\Phi,\ell,L) \Big(\frac{mr\log n}{n}\Big)^{\frac{2\beta}{2\beta+1}} , \label{Thm_bound_4}
	\end{equation}
	where $C_2(a,\Phi,\ell,L) $ is a constant depending on $a,\Phi,\ell,L$.
\end{theorem}

\begin{remark}
When the dimension $m$ degenerates to $1$, bound (\ref{Thm_bound_4}) matches the minimax optimal rate $O(n^{-2\beta/(2\beta+1)})$ for real valued functions over H\"{o}lder class (see \cite{tsybakov2009introduction})
up to some logarithmic factor, which is introduced by the matrix Bernstein inequality, see \cite{tropp2012user}. In section \ref{Lower_bounds}, we show that bound (\ref{Thm_bound_4}) is minimax optimal up to a logarithmic factor. 
\end{remark}

\subsection{Bias reduction through higher order kernels}
If $A(t)$ is not necessarily low rank, we propose an estimator which is easy to implement and prove an upper bound on its risk measured by $L_{\infty}$-norm.
Such estimators are related to another popular approach parallel to local polynomial estimators for bias reduction, namely, using high order kernels to reduce bias. They can also
be applied to another important technique of low rank estimation or approximation via singular value thresholding, see \cite{cai2010singular} and \cite{chatterjee2015matrix}. 
The estimator proposed by \cite{koltchinskii2011nuclear} is shown to be equivalent to soft singular value thresholding of such type of estimators. 

The kernels we are interested in satisfy the following conditions:
\begin{description}
	\item[K1] $K(\cdot)$ is symmetric, i.e. $K(u)=K(-u)$.
	\item[K2] $K(\cdot)$ is compactly supported on $[-1,1]$.
	\item[K3] $R_K := \int^{\infty}_{-\infty} K^2(u) du < \infty$. 
	\item[K4] $K(\cdot)$ is of order $\ell$, where $\ell = \lfloor \beta \rfloor$.
	\item[K5] $K(\cdot)$ is Lipschitz continuous with Lipschitz constant $0<L_K<\infty$.
\end{description}

Consider 
\begin{equation}
\tilde{A}(t) = \frac{m^2}{nh}\sum\limits_{j=1}^n K\Big(\frac{\tau_j-t}{h}\Big)Y_jX_j . \label{Naive_estimator}
\end{equation}
Note that when $K\geq 0$, (\ref{Naive_estimator}) is the solution to the following convex optimization problem
\begin{equation}
\tilde{A}(t) = \arg\min\limits_{S\in \mathbb{D}} \frac{1}{nh}\sum\limits_{j=1}^{n}K\Big(\frac{\tau_j-t}{h}\Big)(Y_j-\langle S,X_j\rangle)^2.
\end{equation}
In the following theorem we prove an upper bound on its global performance measured by $L_{\infty}$-norm over $\Sigma(\beta,L)$. Such kind of bounds is much harder to obtain even for classical matrix lasso problems. The proof of Theorem \ref{Thm_6} can be found in section \ref{ProofThm_6}. 
\begin{theorem}
	\label{Thm_6}
	Under model (\ref{Model}),
	let $(\tau_j, X_j, Y_j)$, $j=1,...,n$ be i.i.d. copies of the random triplet $(\tau,X,Y)$ with
	$X$ uniformly distributed in $\mathcal{X}$, $\tau$ uniformly distributed in $[0,1]$, $X$ and $\tau$ are independent, and 
	$|Y|\leq a$ a.s. for some constant $a>0$;
	let $A$ be any matrix valued function satisfying A1 and A4,
	and kernel $K$ satisfies K1-K5. Denote $\ell=\lfloor\beta\rfloor$.
	Take 
	\begin{equation}
	\tilde{h}_n:=c_*(K)\Big(\frac{a^2 (\ell!)^2 m\log n}{2\beta L^2n}\Big)^{\frac{1}{2\beta+1}},
	\end{equation}
	Then with probability at least $1-n^{-2}$, the estimator defined in (\ref{Naive_estimator}) satisfies
	\begin{equation}
	\sup\limits_{t\in[\tilde{h}_n,1-\tilde{h}_n]}\frac{1}{m^2}\big\|\tilde{A}(t) - A(t)\big\|^2 \leq  C^*(K)\Big(\frac{a^2 (\ell!)^2 m\log n }{2\beta L^2n}\Big)^{\frac{2\beta}{2\beta+1}},  \label{supop}
	\end{equation}
	where $C^*(K)$ and $c_*(K)$ are constants depending on $K$.
\end{theorem}

\begin{remark}
	When $m$ degenerates to $1$, bound (\ref{supop}) coincides with that of real valued case over H\"{o}lder class, which is $O((\frac{\log n}{n})^{2\beta/(2\beta+1)})$, see \cite{tsybakov2009introduction}. 
	In section \ref{Lower_bounds}, we show that bound (\ref{supop}) is minimax optimal up to a logarithmic factor when $m\gg \log n$.
\end{remark}

\section{Lower Bounds Under Matrix Completion Setting}
\label{Lower_bounds}
In this section, we prove the minimax lower bound of estimators (\ref{point_estimator}), (\ref{Global_estimator}) and (\ref{Naive_estimator}). 
In the realm of classical low rank matrix estimation, \cite{negahban2012restricted} studied the optimality issue measured by the Frobenius norm on the classes defined in terms of a "spikeness index" of the true matrix; \cite{rohde2011estimation} derived optimal rates in noisy matrix completion on different classes of matrices for the empirical prediction error; \cite{koltchinskii2011nuclear} established the minimax rates of noisy matrix completion problems up to a logarithmic factor measured by the Frobenius norm. Based on the ideas of \cite{koltchinskii2011nuclear}, standard methods to prove minimax lower bounds in real valued nonparametric estimation in \cite{tsybakov2009introduction}, and some fundamental results in coding theory, we establish the corresponding minimax lower bounds of (\ref{Thm_bound_3}), (\ref{Thm_bound_4}) and (\ref{supop}) which essentially shows that the upper bounds we get are all optimal up to some logarithmic factor.

For the convenience of presentation, we denote by $\inf_{\hat{A}}$ the infimum over all estimators of $A$. We denote by $\mathcal{A}(r,a)$ the set of 
matrix valued functions satisfying A1, A2, A3, and A4. We denote by $\mathcal{P}(r,a)$ the class of distributions of random triplet $(\tau,X,Y)$ that satisfies model (\ref{Model}) with any $A\in \mathcal{A}(r,a)$. 

In the following theorem, we show the minimax lower bound on the pointwise risk. The proof of Theorem \ref{Thm_7} can be found in section \ref{ProofThm_7}.

\begin{theorem}
	\label{Thm_7}
	Under model (\ref{Model}),
	let $(\tau_j, X_j, Y_j)$, $j=1,...,n$ be i.i.d. copies of the random triplet $(\tau,X,Y)$ with
	$X$ uniformly distributed in $\mathcal{X}$, $\tau$ uniformly distributed in $[0,1]$, $X$ and $\tau$ are independent, and 
	$|Y|\leq a$, a.s. for some constant $a>0$;
	let $A$ be any matrix valued function in $\mathcal{A}(r,a)$.
	Then there is an absolute constant $\eta\in (0,1)$ such that for all $t_0\in[0,1]$
	\begin{equation}
	\inf\limits_{\hat{A}}\sup\limits_{P^A_{\tau,X,Y}\in\mathcal{P}(r,a)}\mathbb{P}_{P^A_{\tau,X,Y}}\Big\{\frac{1}{m^2}\big\|\hat{A}(t_0)-A(t_0)\big\|^2_2 > C(\beta,L,a)\Big( \frac{mr}{n}\Big)^{\frac{2\beta}{2\beta+1}} \Big\} \geq \eta. \label{Thm_bound_7}
	\end{equation}
	where $C(\beta,L,a)$ is a constant depending on $\beta$, $L$ and $a$.
\end{theorem}
\begin{remark}
	Note that compared with the upper bound (\ref{Thm_bound_3}), the lower bound  (\ref{Thm_bound_7}) matches it that up to a logarithmic factor. As a consequence, it shows that the estimator (\ref{point_estimator}) achieves a near optimal minimax rate of pointwise estimation. Although, the result of Theorem \ref{Thm_7} is under bounded response condition,  it can be readily extended to the case when the noise in (\ref{Model1}) is Gaussian.
\end{remark}

In the following theorem, we show the minimax lower bound on the integrated risk measured by $L_2$-norm. The proof of Theorem \ref{Thm_8} can be found in section \ref{ProofThm_8}.
\begin{theorem}
	\label{Thm_8}
	Under model (\ref{Model}),
	let $(\tau_j, X_j, Y_j)$, $j=1,...,n$ be i.i.d. copies of the random triplet $(\tau,X,Y)$ with
	$X$ uniformly distributed in $\mathcal{X}$, $\tau$ uniformly distributed in $[0,1]$, $X$ and $\tau$ are independent, and 
	$|Y|\leq a$, a.s. for some constant $a>0$;
	let $A$ be any matrix valued function in $\mathcal{A}(r,a)$.
	Then there is an absolute constant $\eta\in (0,1)$ such that
	\begin{equation}
	\inf\limits_{\hat{A}} \sup\limits_{P^A_{\tau,X,Y}\in\mathcal{P}(r,a)}\mathbb{P}_{P^A_{\tau,X,Y}}\Big\{\frac{1}{m^2}\int^1_0\big\|\hat{A}(t)-A(t)\big\|^2_2dt > \tilde{C}(\beta,L,a)\Big( \frac{mr}{n}\Big)^{\frac{2\beta}{2\beta+1}} \Big\} \geq \eta, \label{Thm_bound_8}
	\end{equation}
	where $\tilde{C}(\beta,L,a)$ is a constant depending on $L$, $\beta$ and $a$.
\end{theorem}
\begin{remark}
	The lower bound in (\ref{Thm_bound_8}) matches the upper bound we get in (\ref{Thm_bound_4}) up to a logarithmic factor. 
	Therefore, it means that the estimator (\ref{Global_estimator}) achieves a near optimal minimax rate on the integrated risk measured by $L_2$-norm. 
	The result of Theorem \ref{Thm_8} can be readily extended to the case when the noise in (\ref{Model1}) is Gaussian.
\end{remark}

Now we consider the minimax lower bound on integrated risk measured by $L_{\infty}$-norm for general matrix valued functions without any rank information. 
Denote
$$
\mathcal{A}(a) := \big\{A(t)\in \mathbb{H}_m,~\forall t\in [0,1]:~|A_{ij}(t)|\leq a,~A_{ij}\in \Sigma(\beta,L)\big\}. \label{aset2}
$$
We denote by $\mathcal{P}(a)$ the class of distributions of random triplet $(\tau,X,Y)$ that satisfies model (\ref{Model}) with any 
$A\in \mathcal{A}(a)$.

In the following theorem, we show the minimax lower bound over $\mathcal{P}(a)$ and $\mathcal{A}(a)$ measured by $L_{\infty}$-norm. The proof of Theorem \ref{Thm_9} can be found in section \ref{ProofThm_9}.
\begin{theorem}
	\label{Thm_9}
	Under model (\ref{Model}),
	let $(\tau_j, X_j, Y_j)$, $j=1,...,n$ be i.i.d. copies of the random triplet $(\tau,X,Y)$ with
	$X$ uniformly distributed in $\mathcal{X}$, $\tau$ uniformly distributed in $[0,1]$, $X$ and $\tau$ are independent, and 
	$|Y|\leq a$, a.s. for some constant $a>0$;
	let $A$ be any matrix valued function in $\mathcal{A}(a) $.
	Then there exist an absolute constant $\eta\in (0,1)$ such that
	\begin{equation}
	\inf\limits_{\hat{A}} \sup\limits_{P^A_{\tau,X,Y}\in\mathcal{P}(a)}\mathbb{P}_{P^A_{\tau,X,Y}}\Big\{\sup\limits_{t\in(0,1)}\frac{1}{m^2}\big\|\hat{A}(t)-A(t)\big\|^2 > \bar{C}(\beta,L,a) \Big( \frac{m \vee \log n}{n}\Big)^{\frac{2\beta}{2\beta+1}} \Big\} \geq \eta. \label{Thm_bound_9}
	\end{equation}
	where $\bar{C}(\beta,L,a)$ is a constant depending on $\beta$, $L$ and $a$.
\end{theorem}
\begin{remark}
	Recall that in the real valued case, the minimax lower bound measured by $L_{\infty}$-norm over H\"{o}lder class is $O((\frac{\log n}{n})^{2\beta/(2\beta+1)})$, see \cite{tsybakov2009introduction}. According to 
	bound (\ref{Thm_bound_9}),
	if dimension $m$ degenerates to $1$, we get the same result as in real valued case and it is optimal. 
	While the dimension $m$ is large enough such that $m\gg \log n$, 
	the lower bound (\ref{Thm_bound_9}) shows that the estimator (\ref{Naive_estimator}) achieves a near optimal minimax optimal rate up to a logarithmic factor.
\end{remark}

\section{Model Selection}
\label{bandwidth}
Despite the fact that estimators (\ref{point_estimator}) and (\ref{Global_estimator}) achieve near optimal minimax rates in theory with properly chosen bandwidth $h$ and order of degree $\ell$, such parameters depend on quantities like $\beta$ and $L$ which are unknown to us in advance. In this section, we propose an adaptive estimation procedure to choose $h$ and $\ell$ adaptively. 

Two popular methods to address such problems are proposed in the past few decades. One is Lepskii's method, and the other is aggregation
method. In the 1990s, many data-driven procedures for selecting the smoothing parameter $h$  emerged. Among them, a series of papers stood out and shaped a method what is now called Lepskii's method. This method has been described in its general form and in great detail in \cite{lepskii1991problem}. Later, \cite{lepski1997optimal} proposed a bandwidth selection procedure based on pointwise adaptation of a kernel estimator that achieves optimal minimax rate of pointwise estimation over H\"{o}lder class, and \cite{lepski1997optimal2} proposed a new
bandwidth selector that achieves optimal rates of convergence over Besov classes with spatially imhomogeneous smoothness.
The basic idea of Lepskii's method is to choose a bandwidth from a geometric grid to get an estimator not very different from those indexed by smaller bandwidths on the grid. 
Although Lepskii's method is shown to give optimal rates in pointwise estimation over H\"{o}lder class in \cite{lepski1997optimal},  
it has a major defect when applied to our problem: the procedure already requires a huge amount of computational cost when real valued functions are replaced by matrix valued functions. Indeed, with Lepskii's method, in order to get a good bandwidth, one needs to compare all candidates indexed by smaller bandwidth with the target one, which leads to dramatically growing computational cost. Still, we have an extra parameter $\ell$ that needs to fit with $h$. 
As a result, we turn to aggregation method to choose a bandwidth from the geometric grid introduced by Lepskii's method, which is more computationally efficient for our problem.
The idea of aggregation method can be briefly summarized as follows: one splits the data set into two parts; the first is used to build all candidate estimators and the second is used to aggregate the estimates to build a new one (aggregation) or select one (model selection) which is at least as good as the best among all candidates. 

The model selection procedure we use was initially introduced by \cite{barron1991complexity} in classical nonparametric estimation with bounded response. 
\cite{wegkamp2003model} generalized this method to the case where the noise can be unbounded but with a finite $p$-th moment for some $p>2$. 
One can find a more detailed review on such penalization methods in \cite{koltchinskii2006local}.

Firstly, we introduce the geometric grid created by \cite{lepski1997optimal} where to conduct our model selection procedure.
Assume that the bandwidth falls into the range $[h_{min}, h_{max}]$.
Recall that the optimal bandwidth $\hat{h}_n$ in theory is given as 
\begin{equation}
\label{band}
\hat{h}_n = C_1 \Big( \frac{ \ell^3\big(\ell !\Phi R(T) a\big)^2 mr\log n }{L^2n} \Big)^{\frac{1}{2\beta+1}}.
\end{equation}
Assume that  $[\beta_*,\beta^*]$ and $[L_*, L^*]$ are the ranges of $\beta,L$ to be considered respectively. Then 
$h_{\max}$ and $h_{\min}$ can be chosen as 
$$
h_{\max} =  C_1 \Big( \frac{\ell ^{*3} \big(\ell ^* !\Phi R(T) a\big)^2 mr\log n }{L_*^2n} \Big)^{\frac{1}{2\beta^*+1}},
$$
and 
$$
h_{\min} =  C_1 \Big( \frac{ \ell _*^3\big( \ell _* !\Phi R(T) a\big)^2  mr\log n}{L^{*2}n} \Big)^{\frac{1}{2\beta_*+1}}
$$
where $\ell^* = \lfloor \beta^*\rfloor$ and $\ell_* = \lfloor \beta_*\rfloor$.
When those ranges are not given, a natural upper bound of $h_{\max}$ is $1$, and a typical choice of $h_{\min}$ can be set to $n^{-1/2}$. 

Denote
$$
d(h) = \sqrt{1\vee 2\log\Big(\frac{h_{max}}{h}\Big)},~~~
d_n = \sqrt{2\log\Big(\frac{h_{\max}}{h_{\min}}\Big)} ,~~~
\alpha(h)=\frac{1}{\sqrt{d(h)}}.
$$
Apparently, $d_n = O(\sqrt{\log n})$. Define grid $\mathcal{H}$ inductively by
\begin{equation}
\mathcal{H} := \Big\{ h_k\in [h_{min}, h_{max}]: h_0=h_{max}, h_{k+1} = \frac{h_k}{1+\alpha(h_k)} ,k=0,1,2,...        \Big\}. \label{grid}
\end{equation}
$\{h_k\}$ on the grid $\mathcal{H}$ is a decreasing sequence and the sequence becomes denser as $k$ grows. 

Now, we consider possible choices of $\ell$. A trivial candidate set is 
 $$
 \mathcal{L}:= \{ \lfloor \beta_* \rfloor, \lfloor \beta_* \rfloor+1,..., \lfloor \beta^* \rfloor\}\subset \NN^*.
 $$ 
If the size of this set is large, one can shrink it through the correspondence (\ref{band}) for each $h_k$. For example, if $n =  \bTheta(m^{d})$ for some $d > 1$, 
one can choose $\ell_i$ such that
$
\big\lfloor \frac{  (1-d^{-1})\log n^{-1} }  {2\log h_k} -0.5  \big\rfloor \leq \ell_i \leq \big\lfloor \frac{\log n^{-1}}{2\log h_k} -0.5 \big\rfloor,
$
which indicates the more the data, the narrower the range. We denote the candidate set for $\ell$ as $\mathcal{L}$. Then the set 
$$
\tilde{\mathcal{H}}=\mathcal{H}\times \mathcal{L}:= \{(h, \ell): h\in\mathcal{H},~\ell \in \mathcal{L}\}
$$
indexed a countable set of candidate estimators. 

\begin{remark}
	In general, selecting $h$ is considered to be more challenging and important than selecting $\ell$ and $\varepsilon$. On one hand, one needs to select $h$ from
	an interval which is an uncountable set compared with selecting $\ell$ from only a finite set of integers. On the other hand, the performance of the estimator is much more sensitive 
	to different choices of $h$, namely, a very small change of $h$ can lead to huge performance degradation. 
	We shall see this through our simulation study in section \ref{section: numerical_model_select}. Once $h$ and $\ell$ are chosen, one can get $\varepsilon_i$ by plug in the value of $(h_i, \ell_i)$
	to get the corresponding
	$\varepsilon_i  = (\ell_i+1) R(T)\Phi \sqrt{\frac{\log2m}{nm h_i }}$.
\end{remark}

Now we introduce our model selection procedure based on $\tilde{\mathcal{H}}$. We split the data $(\tau_j, X_j, Y_j)$, $j=1,...,2n$, into two parts with equal size. The first part of the observations $\{(\tau_j, X_j, Y_j):j\in \hbar_n\}$ contains $n$ data points, which are randomly drawn without replacement from the original 
data set. We construct a sequence of estimators $\hat{A}^k$, $k=1,2,...$ based on the training data set $\hbar_n$ through (\ref{Global_estimator}) for each pair in $\tilde{\mathcal{H}}$.
Our main goal is to select an estimator $\hat{A}$ among $\{\hat{A}^k\}$, which is as good as the one that has the smallest mean square error. We introduce an quantity $\pi_k$ associated with each estimator $\hat{A}^k$ which serves as a penalty term. 
We use the remaining part of the data set $\{(\tau_j,X_j,Y_j):j\in \tau^{\dagger}_n\}$ to perform the selection procedure:
\begin{equation}
k^* = \arg\min\limits_{k}\frac{1}{n}\sum\limits_{j\in \ell_n} (Y_j - \langle \hat{A}^k(\tau_j), X_j\rangle)^2 + \frac{\pi_k}{n}.  \label{Selector}
\end{equation}
Denote $\hat{A}^* = \hat{A}^{k^*}$ as the adaptive estimator. 
In practice, we suggest one to rank all estimators $\hat{A}^k$ according to the following rule: 1. pairs with bigger $h$ always have smaller index; 
2. if two pairs have the same $h$, the one with smaller $\ell$ has smaller index.
Our selection procedure can be summarized in Algorithm \ref{alg1}.

\begin{algorithm}[!htb]
	\begin{enumerate}
		\item Construct the geometric grid $\mathcal{H}$ defined in (\ref{grid}) and the candidate set $\tilde{\mathcal{H}}$;
		\item Equally split the data set $(\tau_j,X_j,Y_j)$, $j=1,...,N$ into two parts $\hbar_n$ and $\tau^{\dagger}_n$ by randomly drawing without replacement;
		\item For each pair in $\tilde{\mathcal{H}}$, construct an estimator $\hat{A}^k$ defined in (\ref{Global_estimator}) using data in $\hbar_n$;
		\item Perform the selection procedure in (\ref{Selector}) using data in $\tau^{\dagger}_n$.
	\end{enumerate}
	\caption{Model Selection Procedure}
	\label{alg1}
\end{algorithm}

The selection procedure described in Algorithm 1 have several advantages: firstly, it chooses a global bandwidth instead of a local one; 
secondly, since our selection procedure as in (\ref{Selector}) is only based on computations of entries of $\hat{A}^k$, no matrix computation is involved in the last step, which can efficiently save computational cost and can be easily applied to high dimensional problems; finally, step $3$ and $4$ can be easily parallelized on distributed platforms. 

The following theorem shows that the integrated risk of $\hat{A}^*$ measured by $L_2$-norm can be bounded by the smallest one among all candidates plus an extra term of order $O(n^{-1})$ which is negligible. 
The proof of Theorem \ref{Thm_10} can be found in section \ref{Proof:Thm_10}.
\begin{theorem}
	\label{Thm_10}
	Under model (\ref{Model}),
	let $(\tau_j, X_j, Y_j)$, $j=1,...,2n$ be i.i.d. copies of the random triplet $(\tau,X,Y)$ with
	$X$ uniformly distributed in $\mathcal{X}$, $\tau$ uniformly distributed in $[0,1]$, $X$ and $\tau$ are independent, and 
	$|Y|\leq a$, a.s. for some constant $a>0$;
	let $A$ be a matrix valued function satisfying A1, A2, A3, and A4;
	let $\{\hat{A}^k\}$ be a sequence of estimators constructed from $\tilde{\mathcal{H}}$;
	let $\hat{A}^*$ be the adaptive estimator selected through Algorithm \ref{alg1}. Then with probability at least $1-n^{-(mr-1)}$
	\begin{equation}
	\frac{1}{m^2}  \int_0^1\big\| \hat{A}^*(t) - A(t)\big\|^2_2dt \leq 3\min\limits_k\Big\{\frac{1}{m^2} \int_0^1\big\| \hat{A}^k(t) - A(t)\big\|^2_2dt + \frac{\pi_k}{n} \Big\} +  \frac{C(a)}{n},
	\end{equation}
	where $C(a)$ is a constant depending on $a$.
\end{theorem}

Recall that ${\rm Card}(\mathcal{H}) = O(\log n)$, one can take $\pi_k = k mr$. Then $\pi_k \leq c_1mr\log n$ uniformly for all $k$ with some numerical constant $c_1$.
According to Lepskii's method that at least one candidate in $\mathcal{H}$ gives the optimal bandwidth associated with the unknown smoothness parameter $\beta$, the following corollary is a direct consequence of Theorem \ref{Thm_4} and \ref{Thm_10}, which shows that $\hat{A}^*$ is adaptive.

\begin{corollary}
	Assume that the conditions of Theorem \ref{Thm_10} hold with $\pi_k = kmr$, and $n > mr\log n$. Then with probability at least $1-n^{-(mr-1)}$
	\begin{equation}
	\frac{1}{m^2}  \int_0^1\big\| \hat{A}^*(t) - A(t)\big\|^2_2dt   \leq C(a,\ell,L) \Big(\frac{mr\log n}{n}\Big)^{\frac{2\beta}{2\beta+1}}
	\end{equation}
	where $C(a, \ell,L)$ is a constant depending on $a$, $\ell$, and $L$.
\end{corollary}

\section{Numerical Simulation}

In this section, we present numerical simulation results of the estimators (\ref{Estimator}) and (\ref{Global_estimator}) to validate the theoretical bounds in (\ref{Thm_bound_3}), (\ref{Thm_bound_4}), 
(\ref{Thm_bound_7}), and (\ref{Thm_bound_8}).
Then we present the simulation results of the model selection procedure shown in Algorithm
\ref{alg1}. Recall that the key optimization problem we need to solve is (\ref{Estimator}). We develop a solver based on the well known alternating direction method of multipliers (ADMM) algorithm \cite{boyd2011distributed} and its applications to matrix recovery problems, see \cite{lin2010augmented,chen2012matrix}. The algorithm can be summarized as in Algorithm \ref{alg:ADMM}.

\begin{algorithm}[!htb]
	\caption{ADMM Algorithm}
	\label{alg:ADMM}
	Set up the values of $max\_Iteration$ and tolerance $\varepsilon_{tol} >0$; \ 
	Initialize $S^{(0)}, \bar{S}^{(0)} \in \DD$ and $Z^{(0)} = \bf{0}$ ; \
	\While {$k < max\_Iteration$}{
		$ S^{(k+1)}  = \arg \min\limits_{S\in \DD}  \frac{1}{nh}\sum\limits_{j=1}^{n}K\Big(\frac{\tau_j-t_0}{h}\Big)\Big(Y_j-\Big\langle \sum\limits_{i=0}^{\ell}S_i p_i\Big(\frac{\tau_j-t_0}{h}\Big),X_j\Big\rangle\Big)^2 
		+ \frac{\rho}{2} \|S-\bar{S}^{(k)}\|_2^2 + \langle Z^{(k)} , S-\bar{S}^{(k)}\rangle$; \
		$ \bar{S}^{(k+1)}  = \arg \min\limits_{\bar{S}\in \DD}  \varepsilon \|\bar{S}\|_1  + \frac{\rho}{2} \|S^{(k+1)}-\bar{S}\|_2^2 + \langle Z^{(k)} , S^{(k+1)}-\bar{S}\rangle;$ \
		$Z^{(k+1)} = Z^{(k)} + \rho (S^{(k+1)}-\bar{S}^{(k+1)})$; \\
		\If{$\|\bar{S}^{(k+1)} - \bar{S}^{(k)}\|_2^2 \leq \varepsilon_{tol}$ or $\|Z^{(k+1)} - Z^{(k)}\|_2^2 \leq \rho^2\varepsilon_{tol}$} {Reaching the tolerance; } {Return $\bar{S}^{(k+1)}$.} \
		$k = k+1;$
	}
	Return $\bar{S}^{(k+1)}$.
\end{algorithm}

The underlying matrix valued function we create is in H\"{o}lder class $\Sigma(\beta, L)$ with $\beta = 3/2$, $L=24$ and rank constraint $r \leq 3$. The orthogonal polynomial we choose is Chebyshev polynomials of the second kind.

\subsection{Simulation results of theoretical bounds}
\begin{figure}[!htb]
	\centering
	\begin{subfigure}[b]{0.23\textwidth}
		\includegraphics[width=\textwidth]{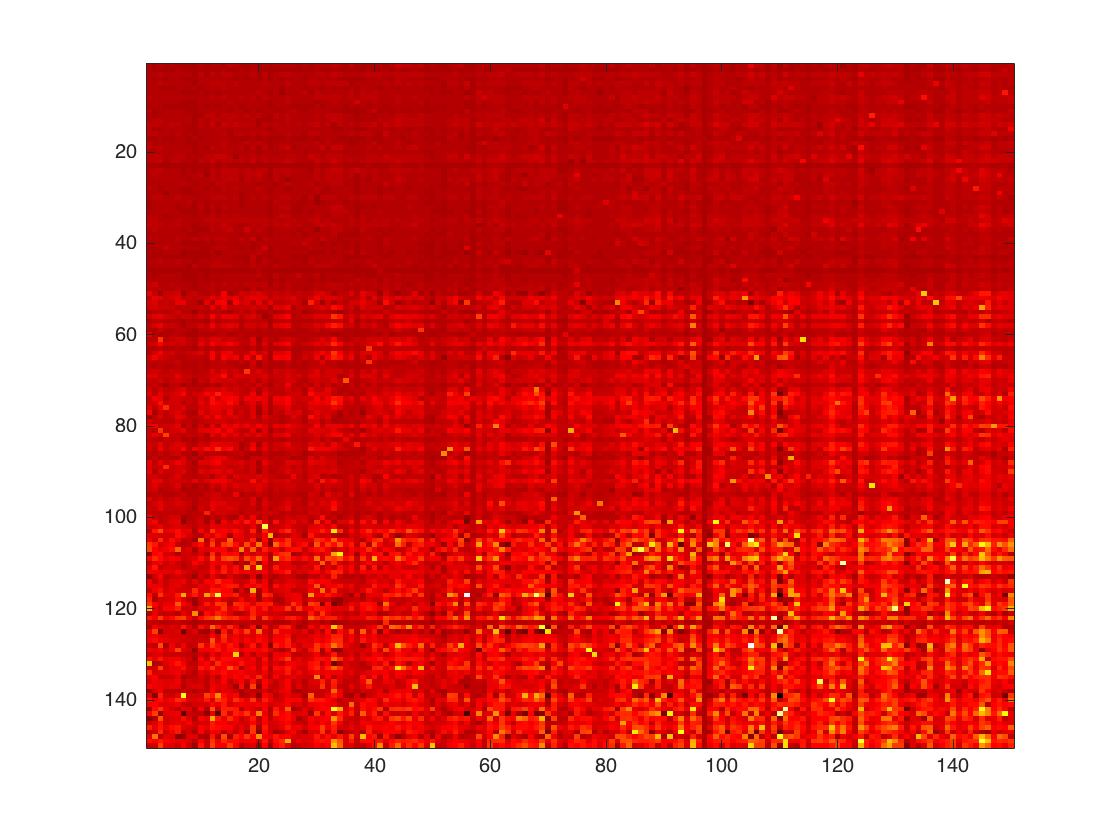}
		\caption{$n=1600$}
		\label{fig:1}
	\end{subfigure}
	~ %add desired spacing between images, e. g. ~, \quad, \qquad, \hfill etc. 
	\begin{subfigure}[b]{0.23\textwidth}
		\includegraphics[width=\textwidth]{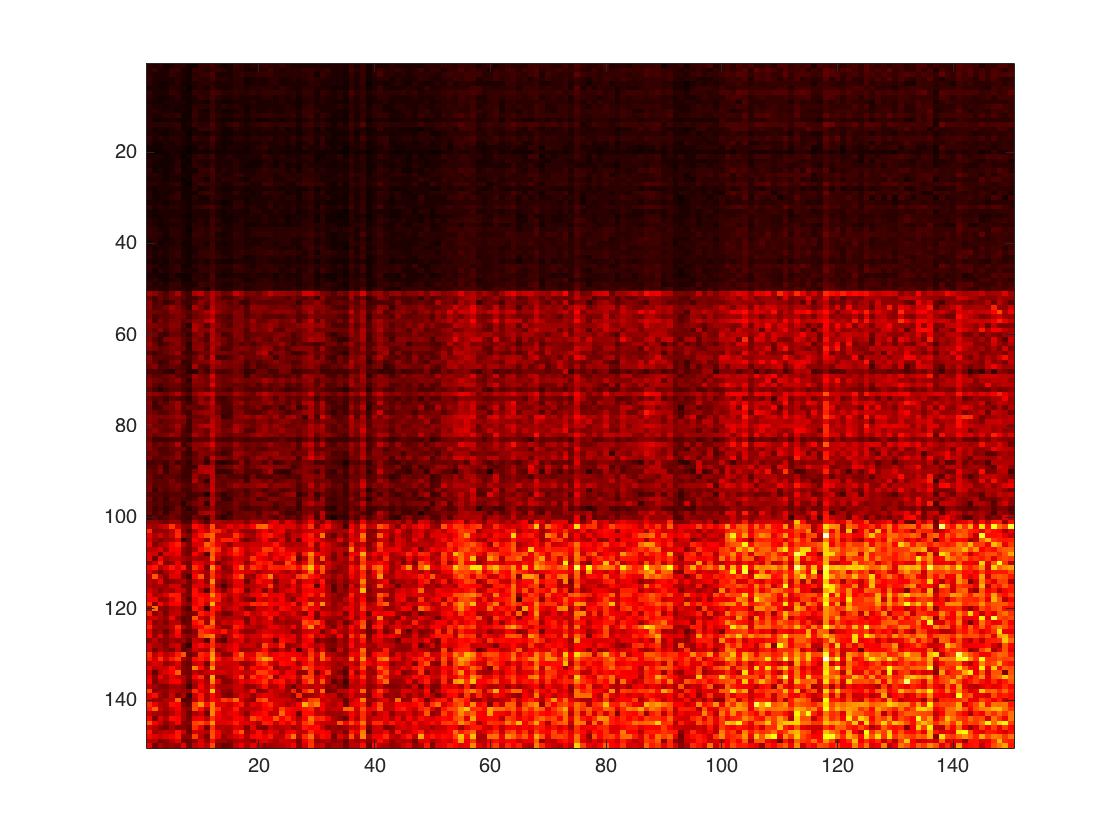}
		\caption{$n=6400$}
		\label{fig:2}
	\end{subfigure}
	~ %add desired spacing between images, e. g. ~, \quad, \qquad, \hfill etc. 
	\begin{subfigure}[b]{0.23\textwidth}
		\includegraphics[width=\textwidth]{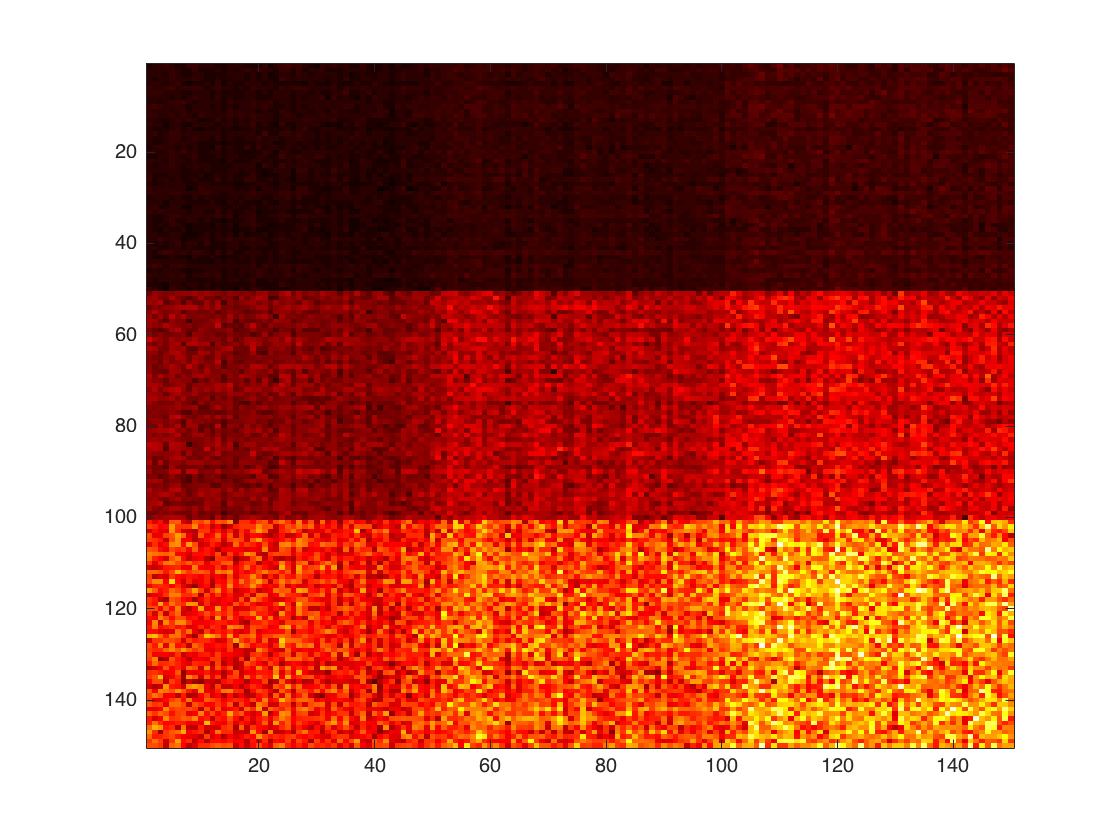}
		\caption{$n=25600$}
				\label{fig:3}
	\end{subfigure}
	\begin{subfigure}[b]{0.23\textwidth}
		\includegraphics[width=\textwidth]{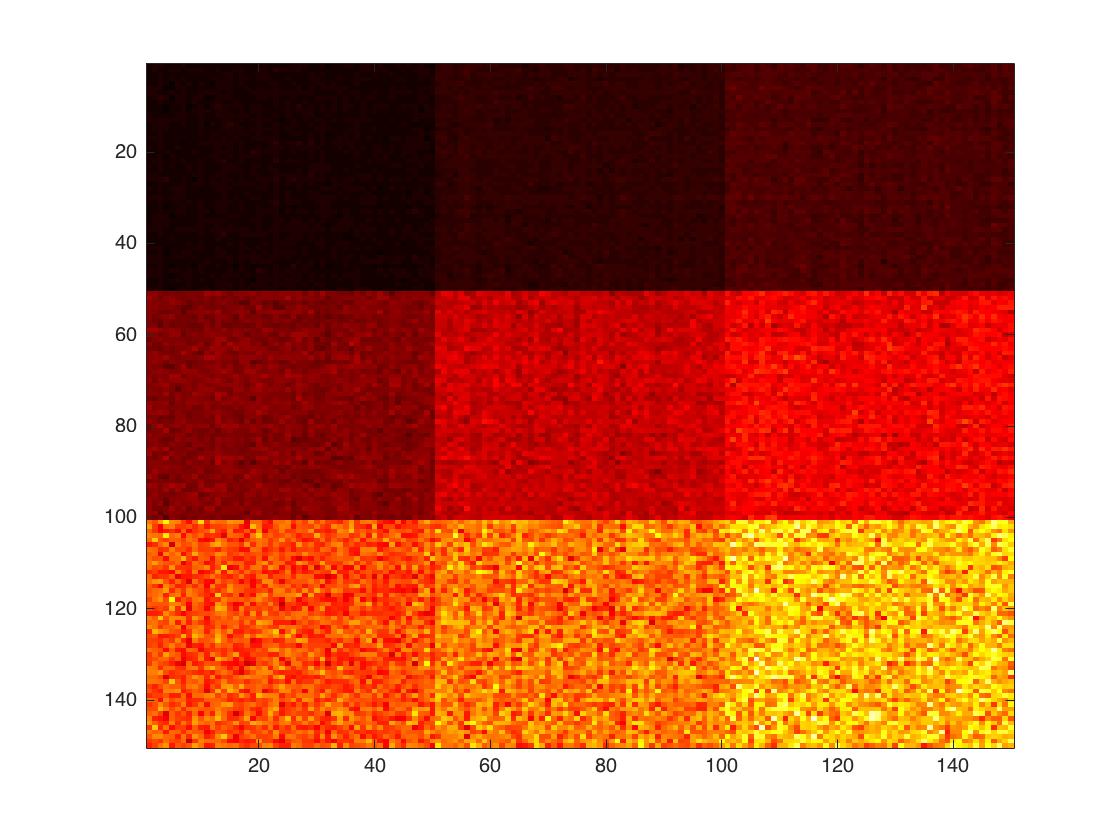}
		\caption{$n=102400$}
				\label{fig:4}
	\end{subfigure}

\centering
\begin{subfigure}[b]{0.23\textwidth}
	\includegraphics[width=\textwidth]{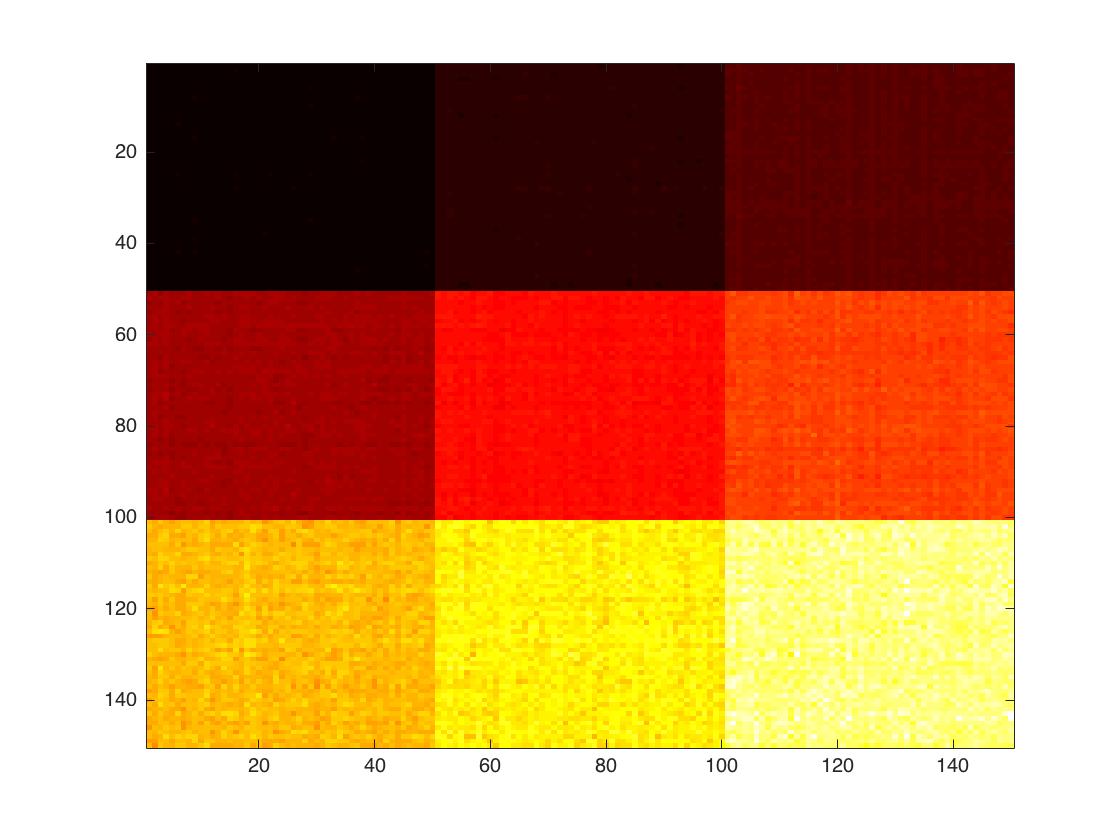}
	\caption{$n=409600$}
		\label{fig:5}
\end{subfigure}
~ %add desired spacing between images, e. g. ~, \quad, \qquad, \hfill etc. 
\begin{subfigure}[b]{0.23\textwidth}
	\includegraphics[width=\textwidth]{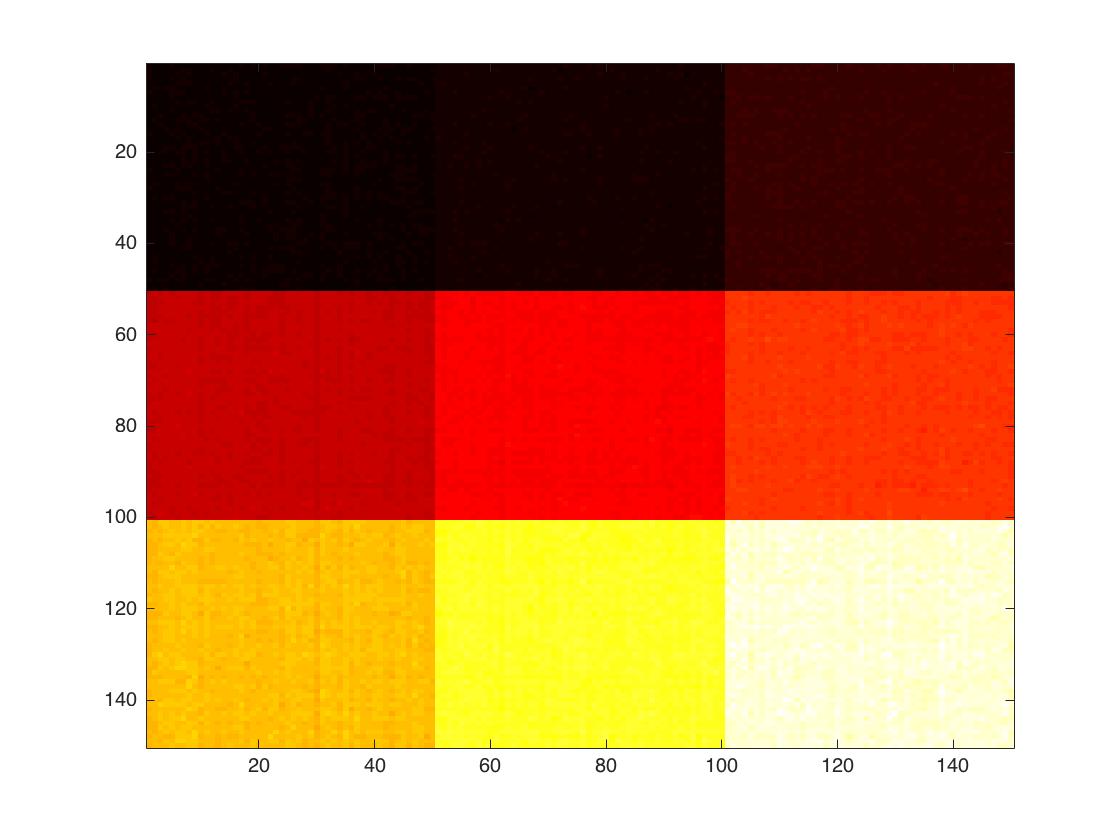}
	\caption{$n=1638400$}
		\label{fig:6}
\end{subfigure}
~ %add desired spacing between images, e. g. ~, \quad, \qquad, \hfill etc. 
\begin{subfigure}[b]{0.23\textwidth}
	\includegraphics[width=\textwidth]{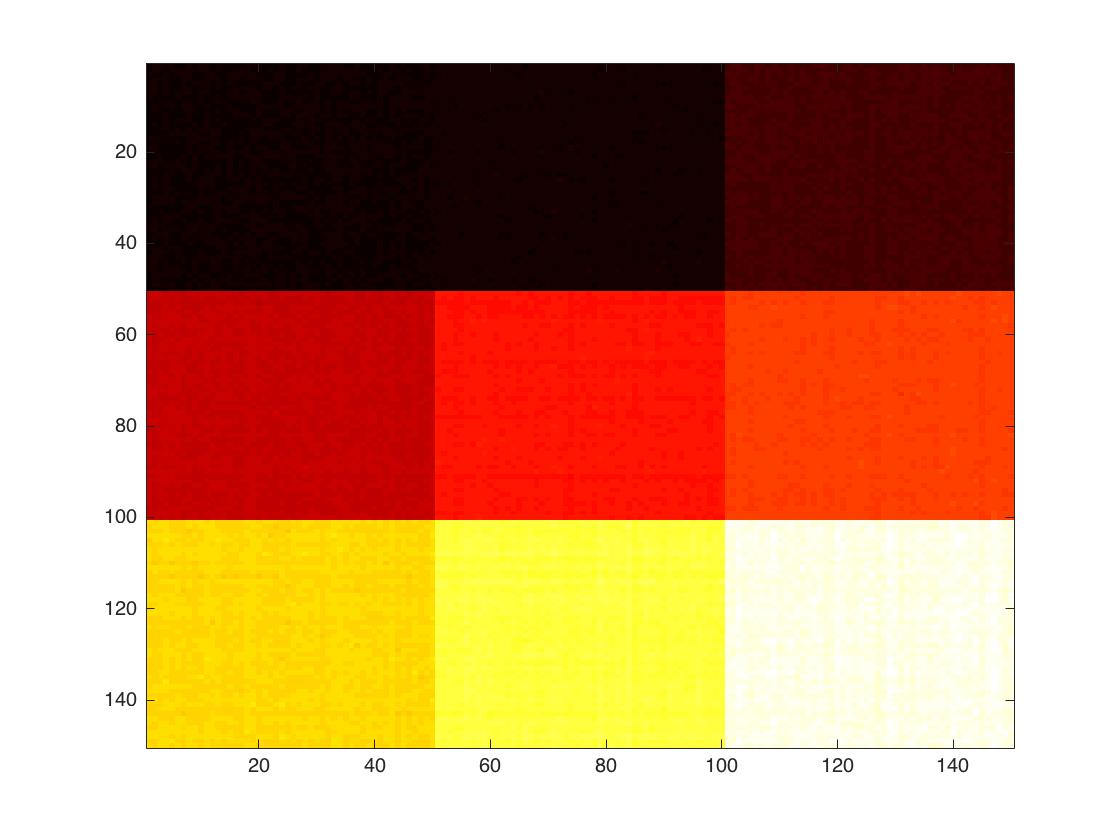}
	\caption{$n=3276800$}
		\label{fig:7}
\end{subfigure}
\begin{subfigure}[b]{0.23\textwidth}
	\includegraphics[width=\textwidth]{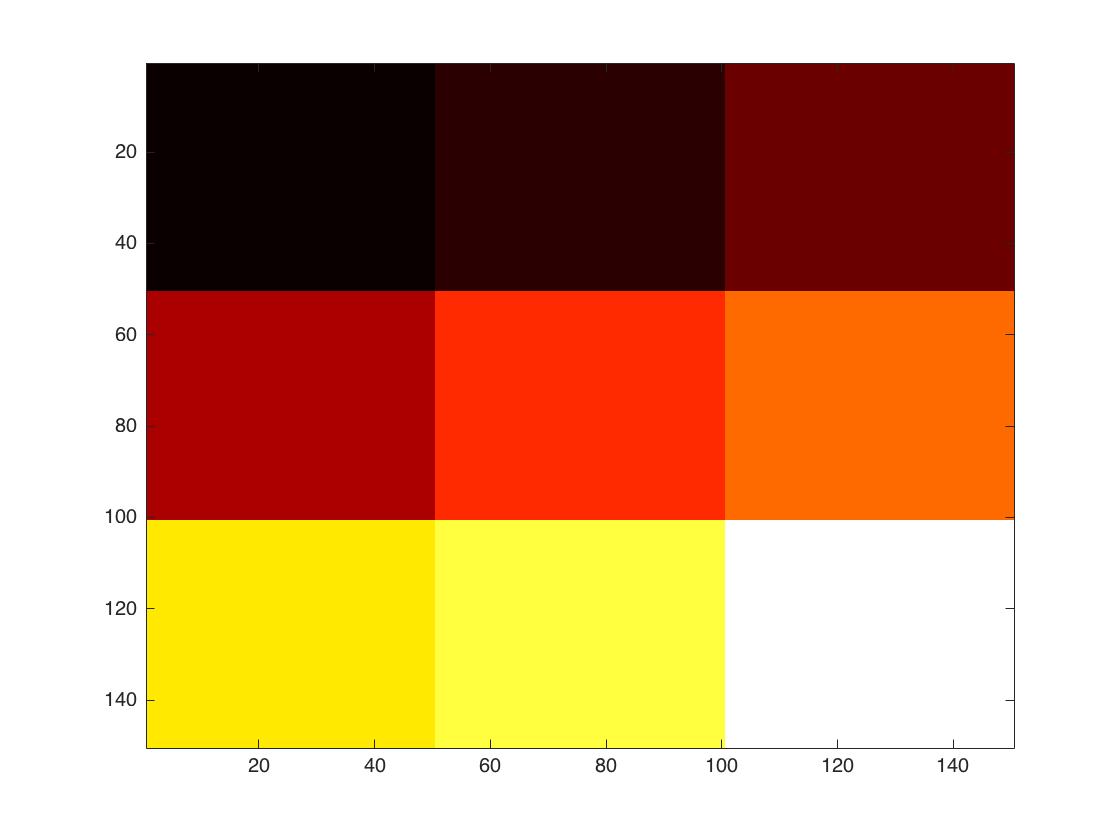}
	\caption{True Data}
		\label{fig:8}
\end{subfigure}
	\caption{Pointwise estimator at $t_0=0.5$ with different sample size}
\end{figure}

We present the numerical simulation results to validates the theoretical bounds that we proved in section \ref{section: local_lasso}, \ref{section: global} and \ref{Lower_bounds}.
By plug in the optimal bandwidth in Theorem \ref{Thm_2}, we run Algorithm \ref{alg:ADMM} to solve the pointwise estimator at $t_0=0.5$ with $m=150$.
Fig. \ref{fig:1} - Fig. \ref{fig:7} show different levels of recovery of the underlying true data matrix as in Fig. \ref{fig:8}. As we can see, the recovery quality increases evidently as sample size $n$ grows. 

\begin{figure}[!htb]
	\centering
	\begin{subfigure}[b]{0.48\textwidth}
		\includegraphics[width=\textwidth]{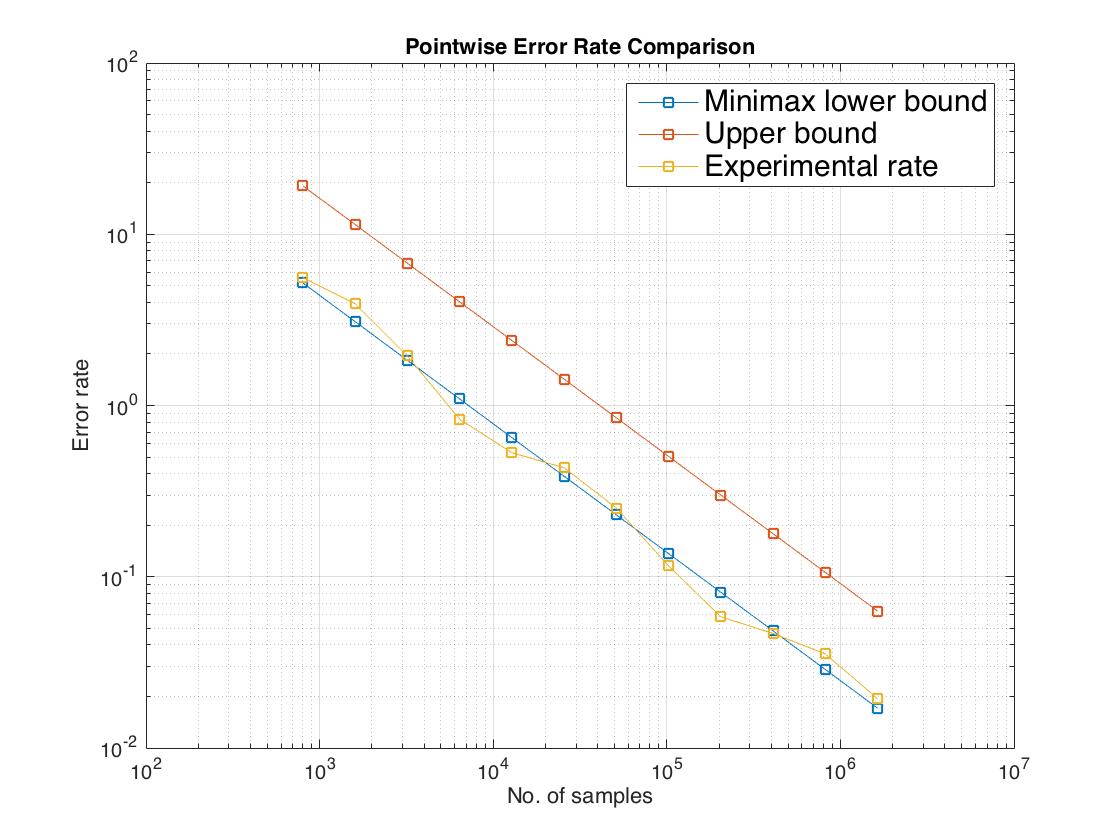}
		\caption{The pointwise risk comparison}
		\label{fig:point_risk}
	\end{subfigure}
	~ %add desired spacing between images, e. g. ~, \quad, \qquad, \hfill etc. 
	\begin{subfigure}[b]{0.48\textwidth}
		\includegraphics[width=\textwidth]{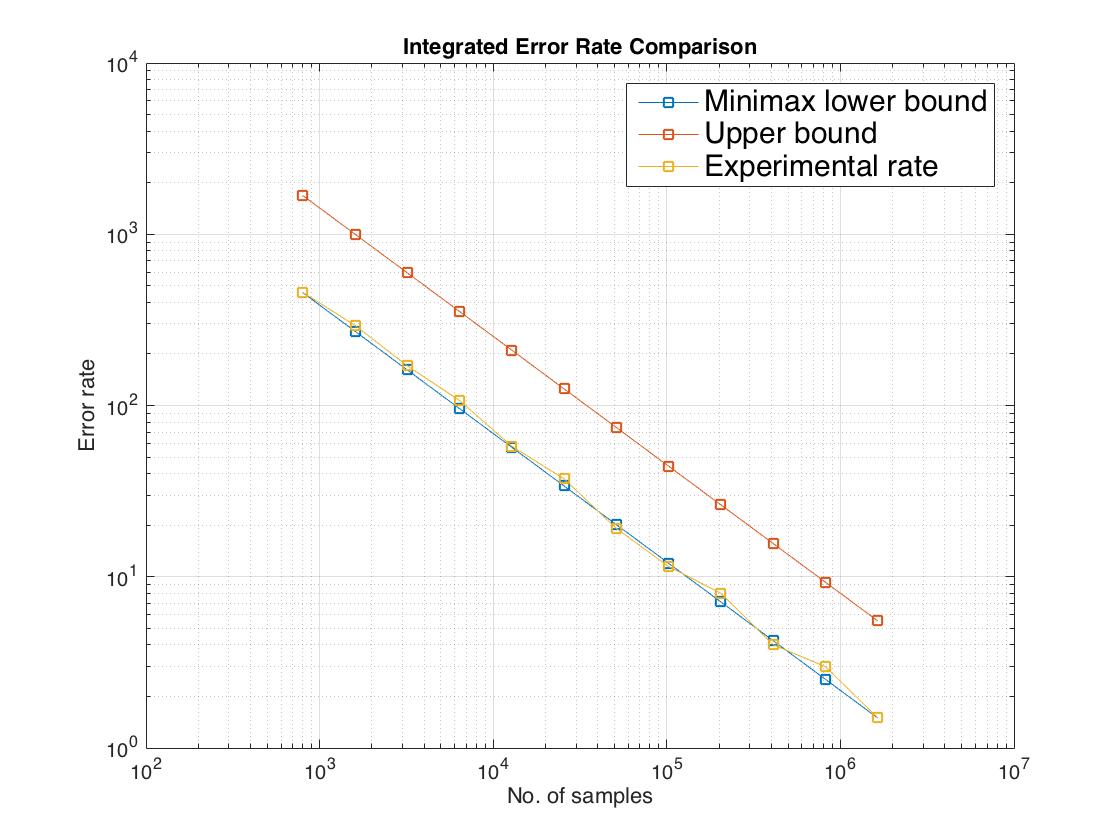}
		\caption{The integrated risk comparison}
		\label{fig:integrated_risk}
	\end{subfigure}
\end{figure}

In Fig. \ref{fig:point_risk}, we display the comparison of pointwise risk between our theoretical bounds proved in 
(\ref{Thm_bound_3}), (\ref{Thm_bound_7}) and our simulation results. In Fig. \ref{fig:integrated_risk}, we display the comparison of integrated risk measured by the $L_2$-norm between the theoretical bounds proved in (\ref{Thm_bound_4}), (\ref{Thm_bound_8}) and our simulation results. Since $\beta = 3/2$ and $\ell = 1$, we use piecewise linear polynomials to approximate the underlying matrix-valued function.
Fig. \ref{fig:point_risk} and \ref{fig:integrated_risk} show that the simulation results match well with the minimax lower bound (\ref{Thm_bound_7}) and (\ref{Thm_bound_8}). One should notice that sometimes our simulated error rate is smaller than the theoretical minimax lower bound. We think the discrepency is due to the fact that the constant factors depending on $a$, $L$ in the minimax lower bound that we computed are not very accurate.

\subsection{Simulation results of model selection}
\label{section: numerical_model_select}
Recall that in section \ref{bandwidth}, we developed Algorithm \ref{alg1} to adaptively choose parameters $h$ and $\ell$.
Since the choice of $\ell$ is made through simply choosing one from a set of integers
and quite straight forward, and choosing a good bandwidth $h$ is more critical and complicated,
we focus on the choice of the smoothing parameter $h$ in our simulation study. We set $\ell = 1$ that is the true parameter.

\begin{figure}[!htb]
	\centering
	\includegraphics[width=0.6\textwidth, height=0.5\textwidth]{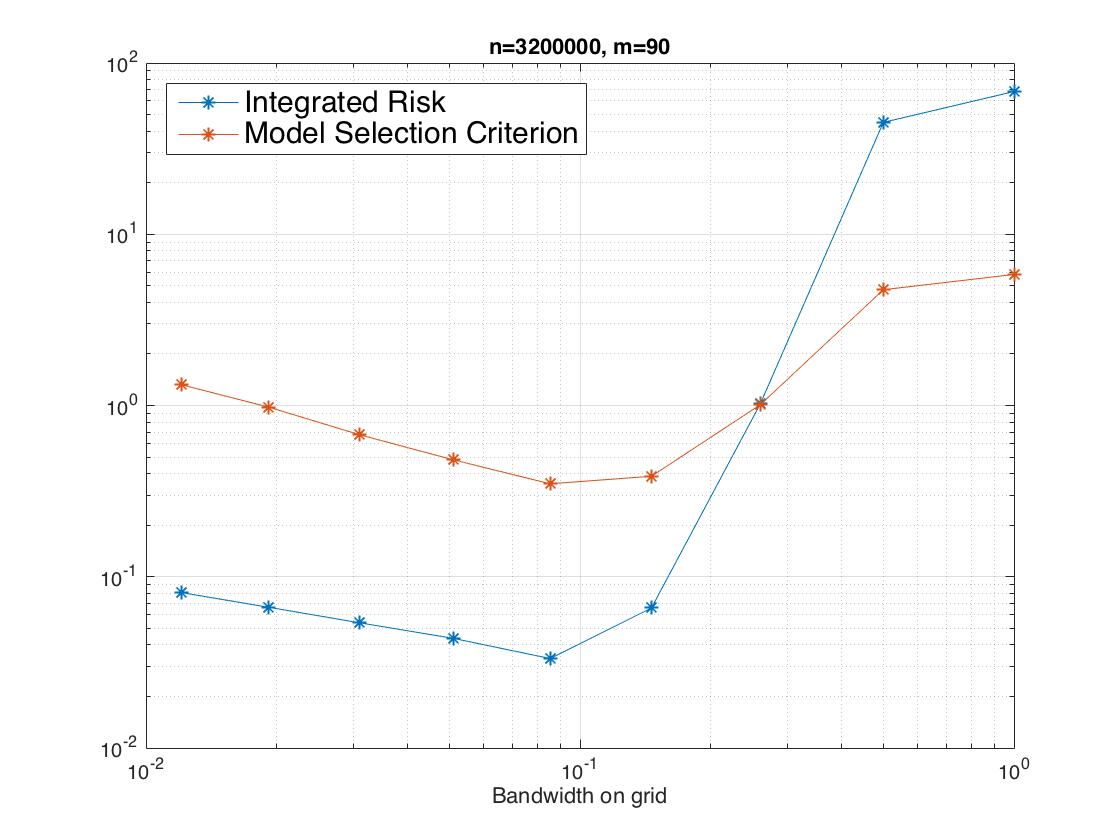}
	\caption{Model Selection on Grid $\mathcal{H}$}
	\label{fig:model_selection}
\end{figure}

We implement Algorithm \ref{alg1} in this section, and perform simulation with $m=90$ and $n=3200000$. 
The theorectially optimal bandwidth $h^*$ is around $0.09$.
We choose $h_{\max} = 1.0$ and $h_{\min} = 0.01$ to construct the geometric grid $\mathcal{H}$ as in (\ref{grid}). 
We display the simulation results in Table \ref{tb:model_select1}. To be more specific, we computed each global estimator as in (\ref{Global_estimator}) with each bandwidth on the $\mathcal{H}$. The corresponding integrated risks measured by $L_2$-norm are displayed in second column and our model selection criterion computed as in (\ref{Selector}) are displayed in the third column. 
One should expect the better integrated risk with smaller value of the third row. The data are plotted in Fig. \ref{fig:model_selection}. 
As we can see, our model selection procedure selects $\hat{h}=0.0853$ with the smallest criterion value of $0.3490$, which shows that $\hat{h}$ is very close to the optimal value of $h$. The corresponding integrated risk is also the smallest among all candidates on the grid and stays very close to the global minimum.

\begin{table}[!htb]
	\centering
	\begin{center}
		\begin{tabular}{|c |c| c| }
			\hline \hline
			Bandwidth on grid $\mathcal{H}$ & Integrated risk  & Model selection criterion \\
			\hline
            1.0000 &  68.1239  &  5.8238  \\
			\hline
			0.5000 &  45.0275  &  4.7442  \\
			\hline
			0.2602 &  1.0207  &  1.0100  \\
			\hline
			0.1461 &  0.0657  &  0.3862  \\
			\hline
			$\mathbf{0.0853}$ &   $\mathbf{0.0333}$   &  $\mathbf{0.3490}$  \\
			\hline
			0.0510 & 0.0437  & 0.4821  \\
			\hline
			0.0311 &  0.0538  &  0.6741 \\
			\hline
			0.0192 & 0.0663  &  0.9771  \\
			\hline
			0.0121 &  0.0807  &  1.3199 \\
			\hline \hline
		\end{tabular}
		\caption{Model Selection}
		\label{tb:model_select1}
	\end{center}
\end{table}

\section{Proofs}
\subsection{Proof of Theorem \ref{Thm_2}}
\label{proof:Thm_2}
\begin{proof}
	Firstly, we introduce a sharp oracle inequality of locally integrated risk of estimator (\ref{local_poly}) in the following lemma. The proof of Lemma \ref{Thm_1} can be found in the appendix, which follows the same 
	derivision as the proof of Theorem 19.1 in \cite{koltchinskii2013sharp}. 
	To be more specific, one just needs to rewrite (\ref{Estimator}) as 
	\begin{equation}
	\label{rewrite}
	\hat{S}^h =\arg\min\limits_{S\in \mathbb{D}} \frac{1}{n}\sum\limits_{j=1}^{n} \Big(\tilde{Y}_j-\Big\langle  S , \tilde{X}_j\Big\rangle\Big)^2 +\varepsilon \|S\|_1. 
	\end{equation}
	where 
	$
	\tilde{X}_j = {\rm Diag}\Big[\sqrt{\frac{1}{h}K\Big(\frac{\tau_j-t_0}{h}\Big)}p_0\Big(\frac{\tau_j-t_0}{h}\Big) X_j,...,\sqrt{\frac{1}{h}K\Big(\frac{\tau_j-t_0}{h}\Big)}p_{\ell}\Big(\frac{\tau_j-t_0}{h}\Big)X_j\Big]
	$,
	and
	$\tilde{Y}_j = \sqrt{\frac{1}{h}K\Big(\frac{\tau_j-t_0}{h}\Big)}Y_j$.	
	Then the proof of Lemma \ref{Thm_1} can be reproduced from the original proof with minor modifications. Since it is mostly tedious repeated arguments, we omit it here. One should notice that in the original proof
	of Theorem 19.1 in \cite{koltchinskii2013sharp}, a matrix isometry condition needs to be satisfied. That is $\|A\|^2_{L_2(\tilde{\Pi})} = \mu_0 \|A\|^2_2$ for some constant $\mu_0 > 0$ and 
	any $A\in \HH_m$ with $\tilde{\Pi}$ being the distribution of $\tilde{X}$. One can easily check that it is true with (\ref{rewrite}). It is also the primary reason why we 
	used orthogonal polynomials instead of the trivial basis $\{1,t,t^2,...\}$.
	\begin{lemma}
		\label{Thm_1}
		Assume that the condition of Theorem \ref{Thm_2} holds.
		Then there exist a numerical constants $D>0$ such that for all
		$$
		\varepsilon \geq D(\ell+1)R(T)\Phi a  \Big(\sqrt{\frac{\log2m}{nmh }} \bigvee \frac{(\log2m)\Phi}{nh}\Big),
		$$
		and for arbitrary $\eta>0$, the estimator (\ref{point_estimator}) satisfies with probability at least $1-e^{-\eta}$
		\begin{equation} 
		\label{ineq:lemma1}
		\begin{aligned}
		&\frac{1}{h}\mathbb{E}K\Big(\frac{\tau-t_0}{h}\Big)\Big\langle A(\tau)- \hat{S}^h(\tau), X\Big\rangle^2\\  
		&\leq \inf\limits_{S\in \mathbb{D}}\Big\{\frac{1}{h}\mathbb{E}K\Big(\frac{\tau-t_0}{h}\Big)\Big\langle A(\tau)-S(\tau), X\Big\rangle^2 \\
		&+ \frac{D^2(\ell+1)^2\Phi^2R(T)^2 a^2 ({\rm rank}(S)m\log 2m+\eta)}{nh} \Big\}.
		\end{aligned}
		\end{equation}
		where $S(\tau):= \sum\limits_{i=0}^{\ell} S_ip_i\big(\frac{\tau-t_0}{h}\big)$.
	\end{lemma}

	Then we consider 
	\begin{equation}
	\label{Thm3:eq1}
	\begin{aligned}
	&\frac{1}{h}\mathbb{E}K\Big(\frac{\tau-t_0}{h}\Big)\Big\langle A(\tau)-\sum\limits_{i=0}^{\ell} \hat{S}^h_ip_i\Big(\frac{\tau-t_0}{h}\Big), X\Big\rangle^2\\
	&=\frac{1}{h}\mathbb{E}K\Big(\frac{\tau-t_0}{h}\Big)\Big\langle A(\tau)-\sum\limits_{i=0}^{\ell} S_ip_i\Big(\frac{\tau-t_0}{h}\Big)+ \sum\limits_{i=0}^{\ell} (S_i-\hat{S}^h_i)p_i\Big(\frac{\tau-t_0}{h}\Big), X\Big\rangle^2\\
	& = \frac{1}{h}\mathbb{E}K\Big(\frac{\tau-t_0}{h}\Big)\Big\langle \sum\limits_{i=0}^{\ell} (S_i-\hat{S}^h_i)p_i\Big(\frac{\tau-t_0}{h}\Big), X\Big\rangle^2  \\
	&+\frac{1}{h}\mathbb{E}K\Big(\frac{\tau-t_0}{h}\Big)\Big\langle A(\tau)-\sum\limits_{i=0}^{\ell} S_ip_i\Big(\frac{\tau-t_0}{h}\Big), X\Big\rangle^2\\
	& + \frac{2}{h}\mathbb{E}K\Big(\frac{\tau-t_0}{h}\Big)\Big\langle A(\tau)-\sum\limits_{i=0}^{\ell} S_ip_i\Big(\frac{\tau-t_0}{h}\Big), X\Big\rangle\Big\langle \sum\limits_{i=0}^{\ell} (S_i-\hat{S}^h_i)p_i\Big(\frac{\tau-t_0}{h}\Big), X\Big\rangle
	\end{aligned}
	\end{equation}
	Therefore, from (\ref{ineq:lemma1}) and (\ref{Thm3:eq1}), we have for any $S\in\DD$
	\begin{equation}
	\label{Thm3:ineq2}
	\begin{aligned}
	&\frac{1}{h}\mathbb{E}K\Big(\frac{\tau-t_0}{h}\Big)\Big\langle \sum\limits_{i=0}^{\ell} (S_i-\hat{S}^h_i)p_i\Big(\frac{\tau-t_0}{h}\Big), X\Big\rangle^2 \\
	& \leq \frac{2}{h}\mathbb{E}K\Big(\frac{\tau-t_0}{h}\Big)\Big|\Big\langle A(\tau)-\sum\limits_{i=0}^{\ell} S_ip_i\Big(\frac{\tau-t_0}{h}\Big), X\Big\rangle\Big\langle \sum\limits_{i=0}^{\ell} (S_i-\hat{S}^h_i)p_i\Big(\frac{\tau-t_0}{h}\Big), X\Big\rangle\Big|\\
	& + \frac{D^2(\ell+1)^2\Phi^2 R(T)^2 a^2 ({\rm rank}(S)m\log 2m+\eta)}{nh} . \\
	& \leq \Big(\frac{c^4}{c^2-1}\Big)\frac{1}{h}\mathbb{E}K\Big(\frac{\tau-t_0}{h}\Big)\Big\langle A(\tau)-\sum\limits_{i=0}^{\ell} S_ip_i\Big(\frac{\tau-t_0}{h}\Big), X\Big\rangle^2\\
	& + \Big(\frac{c^2}{c^2-1}\Big)\Big\{\frac{D^2(\ell+1)^2\Phi^2 R(T)^2a^2 ({\rm rank}(S)m\log 2m+\eta)}{nh}\Big\},
	\end{aligned}
	\end{equation}
	where we used the fact that for any positive constants $a$ and $b$, $2ab \leq \frac{1}{c^2}a^2+c^2b^2$ for some $c>1$.
	Take $S$ such that
	\begin{equation}
	\sum\limits_{i=0}^{\ell} S_ip_i\Big(\frac{\tau-t_0}{h}\Big) = A(t_0) + A'(t_0)h\Big(\frac{\tau-t_0}{h}\Big)+...+ \frac{A^{(\ell)}(t_0)h^{\ell}}{\ell !}\Big(\frac{\tau - t_0}{h}\Big)^{\ell}.  \label{ProofThm_2_1}
	\end{equation}
	Note that this is possible since the right hand side is a matrix valued polynomial of $\frac{\tau-t_0}{h}$ up to order $\ell$, and ${\rm span}\{p_0(t),p_1(t),...,p_{\ell}(t)\}={\rm span}\{1,t,...,t^{\ell}\}$.
	Under the condition that all entries of $A^{(k)}(t)$ are bounded by $a$, then entries of $S_k$ are bounded by $R(T)a$. Thus, the corresponding $S\in\DD$.
	Obviously, ${\rm rank}(S_i)\leq (\ell+1-i) r$.
	Since $A\in\Sigma(\beta,L)$, we consider $\ell$-th order Taylor expansion of $A$ at $t_0$ to get
	\begin{equation}
	A(\tau) = A(t_0) + A'(t_0)(\tau-t_0)+...+ \frac{\tilde{A}(\tau - t_0)^{\ell}}{\ell!}, \label{ProofThm_2_2}
	\end{equation}
	where $\tilde{A}$ is the matrix with $\tilde{A}_{ij} = A^{(\ell)}_{ij}(t_0+\alpha_{ij}(\tau-t_0))$ for some $\alpha_{ij}\in [0,1]$. 
	Then we apply the Taylor expansion (\ref{ProofThm_2_2}) and identity (\ref{ProofThm_2_1}) to get
	\begin{equation}
	\label{Thm3:ineq5}
	\begin{aligned}
	&\frac{1}{h}\mathbb{E}K\Big(\frac{\tau-t_0}{h}\Big)\Big\langle A(\tau)-\sum\limits_{i=0}^{\ell} S_ip_i\Big(\frac{\tau-t_0}{h}\Big), X\Big\rangle^2\\
	&\leq \frac{1}{h}\mathbb{E}K\Big(\frac{\tau-t_0}{h}\Big) \frac{1}{m^2}\Big\| \frac{LU(\tau - t)^{\beta}}{\ell!}\Big\|_2^2\leq  \frac{L^2 h^{2\beta} }{(\ell !)^2}.
	\end{aligned}
	\end{equation}
	where $U$ denotes the matrix with all entries being $1$. The first inequality is due to $A_{ij} \in \Sigma(\beta,L)$, and the second is due to $|\tau-t_0|\leq h$.
	Under the condition that $X$ is uniformly distributed in $\mathcal{X}$, and the orthogonality of $\{p_i(t)\}_{i=0}^{\ell}$, it is easy to check that 
	\begin{equation}
	\label{Thm3:ineq3}
	\frac{1}{h}\mathbb{E}K\Big(\frac{\tau-t_0}{h}\Big)\Big\langle \sum\limits_{i=0}^{\ell} (S_i-\hat{S}^h_i)p_i\Big(\frac{\tau-t_0}{h}\Big), X\Big\rangle^2 = \frac{1}{m^2}\sum\limits_{i=0}^{\ell} \|\hat{S}^h_i-S_i\|_2^2 
	\end{equation}
	Note that 
	\begin{equation}
	\label{Thm3:ineq4}
	\big\|\hat{S}^h(t_0)-S(t_0)\big\|_2^2 = \big\|\sum\limits_{i=0}^{\ell} (\hat{S}^h_i-S_i)p_i(0)\big\|^2_2 \leq (\ell+1) \Phi^2 \sum\limits_{i=0}^{\ell} \big\|\hat{S}^h_i-S_i\big\|_2^2,
	\end{equation}
	where the second inequality is due to Cauchy-Schwarz inequality and $p_i(t)$ are uniformly bounded on [-1,1].
	Combining (\ref{Thm3:ineq2}), (\ref{Thm3:ineq5}), (\ref{Thm3:ineq3}), and (\ref{Thm3:ineq4}),
	we get with probability at least $1-e^{-\eta}$
	\begin{align*}
	&\frac{1}{m^2} \|\hat{S}^h(t_0) - A(t_0)\|_2^2 \leq \Big(\frac{c^4}{c^2-1}\Big) \frac{2L^2 h^{2\beta} }{(\ell !)^2} \\
	&+ \Big(\frac{c^2}{c^2-1}\Big)\Big\{\frac{D^2(\ell+1)^2\Phi^2R(T)^2 a^2 ({\rm rank}(S)m\log 2m+\eta)}{nh}\Big\}.
	\end{align*}
	By optimizing the right hand side with respect to $h$ and take $\eta = mr\log n$, we take 
	$$
	\hat{h}_n = C \Big( \frac{\ell^3 (\ell !)^2\Phi^2 R(T)^2 a^2 mr\log n }{L^2n} \Big)^{\frac{1}{2\beta+1}},
	$$
	where $C$ is a numerical constant.
	This completes the proof of the theorem.
\end{proof}

\subsection{Proof of Theorem \ref{Thm_4}}
\label{ProofThm_4}
\begin{proof}
	It is easy to see that
	\begin{equation}
	\label{Thm_4:ineq1}
	\int^1_0 \|\hat{A}(t)- A(t)\|_2^2 dt \leq \sum\limits_{k=1}^{M/2} \int^{t_{2k-1}+\hat{h}_n}_{t_{2k-1}-\hat{h}_n} \|\hat{S}_k^h(t)- A(t)\|_2^2 dt.
	\end{equation}
	For each $k$, 
	\begin{align*}
	\frac{1}{m^2} \int^{t_{2k-1} + \hat{h}_n}_{t_{2k-1} - \hat{h}_n} &\big\|\hat{S}_k^h(t)- A(t)\big\|_2^2dt \\
	&= \EE_{\tau,X} \mathbf{1}\{(t_{2k-1}-\hat{h}_n, t_{2k-1}+\hat{h}_n]\} \Big\langle A(\tau)- \hat{S}^h(\tau), X\Big\rangle^2
	\end{align*}
	By (\ref{ineq:lemma1}), (\ref{Thm3:ineq5}) and arguments used to prove Theorem \ref{Thm_2}, we have with probability at least $1-\frac{1}{ n^{mr}}$,
	$$
	\frac{1}{m^2 \hat{h}_n} \int^{t_{2k-1} + \hat{h}_n}_{t_{2k-1} - \hat{h}_n} \big\|\hat{S}_k^h(t)- A(t)\big\|_2^2dt \leq C_1(a,\Phi,\ell,L) \Big(\frac{mr\log n}{n}\Big)^{\frac{2\beta}{2\beta+1}}.
	$$
	Then take the union bound over $k$, from (\ref{Thm_4:ineq1}) we get with probability at least $1-\frac{1}{ n^{mr-1}}$,
	$$
	\frac{1}{m^2}\int^1_0 \|\hat{A}(t)- A(t)\|_2^2 dt \leq C_2(a,\Phi,\ell,L) \Big(\frac{mr\log n}{n}\Big)^{\frac{2\beta}{2\beta+1}}.
	$$
	where $C_2(a,\Phi,\ell,L) $ is a constant depending on $a,\Phi,\ell,L$.
\end{proof}

\subsection{Proof of Theorem \ref{Thm_6}}
\label{ProofThm_6}
\begin{proof}
	In this proof, we use $C(K)$ to denote any constant depending on $K$ which may vary from place to place. This simplifies the presentation while does no harm to the soundness of our proof.
	
	Consider 
	\begin{equation}
	\sup\limits_{t\in[\tilde{h}_n,1-\tilde{h}_n]} \|\tilde{A}(t) - A(t)\| \leq \sup\limits_{t\in[\tilde{h}_n,1-\tilde{h}_n]} \|\tilde{A}(t) - \mathbb{E}\tilde{A}(t)\| + \sup\limits_{t\in[\tilde{h}_n,1-\tilde{h}_n]} \|\mathbb{E}\tilde{A}(t) - A(t)\|.
	\end{equation}
	The first term on the right hand side is recognized as the variance and the second is the bias. Firstly, we deal with the bias term.
	Denote $B(t_0) := \mathbb{E}\tilde{A}(t_0) - A(t_0)$, $t_0\in[\tilde{h}_n,1-\tilde{h}_n]$. 
	Recall from (\ref{Model1}), $\mathbb{E}(\xi_j | \tau_j, X_j) = 0$ for any $t_0\in [\tilde{h}_n,1-\tilde{h}_n]$ we have 
	\begin{align*}
	\mathbb{E}\tilde{A}(t_0) &= \mathbb{E}\frac{m^2}{nh}\sum\limits_{j=1}^n K\Big(\frac{\tau_j-t_0}{h}\Big)(\langle A(\tau_j),X_j\rangle + \xi_j)X_j \\
	&= \frac{m^2}{h} \mathbb{E}K\Big(\frac{\tau-t_0}{h}\Big)\langle A(\tau),X\rangle X.
	\end{align*}
	By applying the Taylor expansion of $A(\tau)$ as in (\ref{ProofThm_2_2}) and the fact that $K$ is a kernel of order $\ell$, we get
	$$
	\mathbb{E}\tilde{A}(t_0) = \mathbb{E}\frac{m^2}{h}K\Big(\frac{\tau-t_0}{h}\Big) \langle A(t_0), X\rangle X + \mathbb{E}\frac{m^2}{h}K\Big(\frac{\tau-t_0}{h}\Big) \frac{(\tau-t_0)^{\ell}}{\ell!}\langle \tilde{A}, X\rangle X,
	$$
	where $\tilde{A}$ is the same as in (\ref{ProofThm_2_2}).
	It is easy to check that the first term on the right hand side is $A(t_0)$. Therefore we rewrite $B(t_0)$ as
	\begin{align*}
	B(t_0) &= \mathbb{E}\frac{m^2}{h}K\Big(\frac{\tau-t_0}{h}\Big) \frac{(\tau-t_0)^{\ell}}{\ell!}\langle \tilde{A}, X\rangle X\\
	& = \mathbb{E}\frac{m^2}{h}K\Big(\frac{\tau-t_0}{h}\Big) \frac{(\tau-t_0)^{\ell}}{\ell!}\langle \tilde{A}- A^{(\ell)}(t_0), X\rangle X,
	\end{align*}
	where the second equity is due to the fact that each element of $A(t)$ is in $\Sigma(\beta, L)$ and $K$ is a kernel of order $\ell$.  Then we can bound each element of matrix $B(t_0)$ as
	$$
	\begin{aligned}
	|B_{ij}(t_0)| &\leq \int_0^1 \frac{1}{h} K\Big(\frac{\tau-t_0}{h}\Big) \frac{|\tau-t_0|^{\ell}}{\ell!} |a_{ij}^{(\ell)}(t_0+\alpha(\tau-t_0))-a_{ij}^{(\ell)}(t_0)|d\tau \\
	&\leq  L \int_0^1 |K(u)| \frac{|uh|^{\beta}}{\ell!}du\\
	&\leq C(K) \frac{Lh^{\beta}}{\ell!}.
	\end{aligned}
	$$
	Thus
	\begin{equation}
	\sup\limits_{t\in[\tilde{h}_n,1-\tilde{h}_n]}\|B(t)\| \leq C(K) \frac{Lmh^{\beta}}{\ell!}. \label{bias}
	\end{equation}
	
	On the other hand, for the variance term $\sup_{t\in[\tilde{h}_n,1-\tilde{h}_n]} \|\tilde{A}(t) - \mathbb{E}\tilde{A}(t)\|^2 $, we construct a $\delta-net$ on the interval $[0,1]$ with $\delta = 1/M$, and
	$$
	M =n^2,~~~~t_j = \frac{2j-1}{2M},~j=1,...,M.
	$$ 
	Denote 
	$
	S_n(t) := \tilde{A}(t) - \mathbb{E}\tilde{A}(t),
	$
	then we have
	\begin{equation}
	\sup\limits_{t\in[\tilde{h}_n,1-\tilde{h}_n]} \|S_n(t)\| \leq \sup\limits_{t\in[0,1]} \|S_n(t)\| \leq \max\limits_{i} \|S_n(t_i)\| + \sup\limits_{|t-t'|\leq \delta} \|S_n(t)- S_n(t')\|. \label{ProofThm_6_1}
	\end{equation}
	Next, we bound both terms on the right hand side respectively.
	For each $t_i$, 
	$$
	\begin{aligned}
	S_n(t_i) &= \frac{m^2}{nh}\sum\limits_{j=1}^{n} \Big(K\Big(\frac{\tau_j-t_i}{h}\Big)Y_j X_j -\mathbb{E} K\Big(\frac{\tau_j-t_i}{h}\Big)Y_j X_j\Big).
	\end{aligned}
	$$
	The right hand side is a sum of zero mean random matrices, we apply the matrix Bernstein inequality, see \cite{tropp2012user}. Under the assumption of Theorem \ref{Thm_6},
	one can easily check that with probability at least $1-e^{-\eta}$,
	$$
	\|S_n(t_i)\| \leq C(K)m^2\Big( \sqrt{\frac{a^2(\eta+\log 2m)}{mnh}} \bigvee \frac{a(\eta+ \log 2m)}{nh} \Big).
	$$
	Indeed, by setting $\bar{X} = m^2/hK\big(\frac{\tau-t}{h}\big)YX- \EE  m^2/hK\big(\frac{\tau-t}{h}\big)YX$, it is easy to check that $U_{\bar{X}} \lesssim \|K\|_{\infty}am^2/h$ and $\sigma^2_{\bar{X}} \lesssim R_K a^2 m^3 /h $.
	By taking the union bound over all $i$ and setting $\eta = 4\log n$, we get with probability at least $1-n^{-2}$, 
	$$
	\max\limits_{i} \big\|S_n(t_i)\big\|^2 \leq C(K) \frac{a^2m^3\log n}{nh}.
	$$
	
	As for the second term on the right hand side of (\ref{ProofThm_6_1}), by the assumption that $K$ is a Lipschitz function with Lipschitz constant $L_K$, we have
	$$
	\begin{aligned}
	\sup\limits_{|t-t'|\leq \delta} \|S_n(t)- S_n(t')\| &\leq \sup\limits_{|t-t'|\leq \delta} \|(\tilde{A}(t)- \tilde{A}(t'))\| +  \sup\limits_{|t-t'|\leq \delta}\|\mathbb{E}(\tilde{A}(t)-\tilde{A}(t'))\|\\
	& \leq \frac{L_K a m^3}{n^2h^2} + \frac{L_K a m}{n^2h^2}.
	\end{aligned}
	$$
	Thus with probability at least $1-n^{-2}$, 
	$$
	\sup\limits_{t\in[\tilde{h}_n,1-\tilde{h}_n]}\|S_n(t)\| ^2 \leq C(K) \frac{a^2m^3\log n}{nh}
	$$
	Together with the upper bound we get on the bias in (\ref{bias}), we have with probability at least $1-n^{-2}$, 
	$$
	\sup\limits_{t\in[\tilde{h}_n,1-\tilde{h}_n]}\frac{1}{m^2}\|\tilde{A}(t)- A(t)\| ^2 \leq C(K)\Big( \frac{a^2m\log n}{nh}+ \frac{L^2h^{2\beta}}{\ell!^2}\Big).
	$$
	Choose
	$$
	\tilde{h}_n = C(K)\Big(\frac{a^2(\ell!)^2m\log n}{2\beta L^2n}\Big)^{\frac{1}{2\beta+1}},
	$$
	we get
	$$
	\sup\limits_{t\in[\tilde{h}_n,1-\tilde{h}_n]} \frac{1}{m^2}\|\tilde{A}(t)- A(t)\| ^2 \leq C(K) \Big(\frac{a^2 (\ell!)^2 m\log n}{2\beta L^2n}\Big)^{\frac{2\beta}{2\beta+1}}.
	$$
\end{proof}

\subsection{Proof of Theorem \ref{Thm_7}}
\label{ProofThm_7}
\begin{proof}
	Without loss of generality, we assume that both $m$ and $r$ 
	are even numbers. We introduce several notations which are key to construct the hypothesis set.
	For some constant $\gamma >0 $, denote
	$$
	\mathcal{C}=\big\{\tilde{A}=(a_{ij})\in \CC^{\frac{m}{2}\times \frac{r}{2}}: a_{ij}\in\{0,\gamma\},\forall 1\leq i\leq m/2, 1\leq j \leq r/2 \big\},
	$$
	and consider the set of block matrices
	\begin{equation}
	\mathcal{B}(\mathcal{C})
	= \Bigg\{
	\begin{bmatrix}
	\tilde{A} & \tilde{A} & \dots & \tilde{A} & O \\
	\end{bmatrix}
	\in \CC^{\frac{m}{2}\times \frac{m}{2}}: \tilde{A}\in \mathcal{C}
	\Bigg\},
	\end{equation}
	where $O$ denotes the $ m/2\times (m/2-r\lfloor m/r\rfloor/2)$ zero matrix.
	Then we consider a subset of Hermitian matrices $\mathcal{S}_m\subset \mathbb{H}_m$,
	\begin{equation}
	\mathcal{S}_m
	= \Bigg\{
	\begin{bmatrix}
	\tilde{O} & \hat{A} \\
	\hat{A}^* & \tilde{O}
	\end{bmatrix}
	\in \mathbb{C}^{m \times m}: \hat{A}\in \mathcal{B}(\mathcal{C})
	\Bigg\}.
	\end{equation}
	An immediate observation is that for any matrix $A\in \mathcal{S}_m$, ${\rm rank}(A)\leq r$.
	
	Due to the Varshamov-Gilbert bound (see
	Lemma 2.9 in \cite{tsybakov2009introduction}), there exists a subset $\mathcal{A}^0 \subset \mathcal{S}_m$ with cardinality ${\rm Card}(\mathcal{A}^0)\geq 2^{mr/32} + 1$ containing the zero $m\times m$
	matrix $\mathbf{0}$ such that for any two distinct elements $A_1$ and $A_2$ of $\mathcal{A}^0$,
	\begin{equation}
	\|A_1-A_2\|_2^2 \geq \frac{mr}{16}\lfloor \frac{m}{r}\rfloor \gamma^2 \geq \gamma^2\frac{m^2}{32}. \label{ProofThm_7_1}
	\end{equation}
	
	Let $f_n(t)$ denote the function
	$
	f_n(t):=L h_n ^{\beta} f\big(\frac{t-t_0}{h_n}\big),~t\in[0,1],
	$
	where $h_n= c_0 \big(\frac{mr}{n}\big)^{1/(2\beta+1)}$, with some constant $c_0>0$,
	and $f\in \Sigma(\beta,1/2) \cap C^{\infty}~and ~~Supp(f)=[-1/2,1/2].$
	Note that there exist functions $f$ satisfying this condition. For instance, one can take 
	\begin{equation}
	f(t) = \alpha e^{-\frac{1}{1-4u^2}}\mathbb{I}(|u|<1/2), \label{ProofThm_7_5}
	\end{equation}
	for some sufficient small $\alpha >0$.
	It is easy to check that
	$f_n(t)\in \Sigma(\beta,L)$ on $[0,1]$.
	
	We consider the following hypotheses of $A$ at $t_0$:
	$$
	\mathcal{A}^{\beta}_0:=\big\{\hat{A}(t)=Af_n(t), t\in[0,1]: A\in \mathcal{A}^0\big\}.
	$$
	The following claims are easy to check:
	firstly, any element in $\mathcal{A}^{\beta}_0$ together with its derivative have rank uniformly bounded by $r$, and the difference of any two elements of $\mathcal{A}^{\beta}_0$ satisfies the same property for fixed $t_0$;
	secondly, the
	entries of any element of $\mathcal{A}^{\beta}_0$ together with its derivative are uniformly bounded by some constant for sufficiently small chosen $\gamma$;
	finally, each element of $A(t)\in \mathcal{A}^{\beta}_0$ belongs to $\Sigma(\beta,L)$. Therefore, $\mathcal{A}^{\beta}_0\subset \mathcal{A}(r,a)$ with some chosen $\gamma$.
	
	According to (\ref{ProofThm_7_1}), for any two distinct elements $\hat{A}_1(t)$ and $\hat{A}_2(t)$ of $\mathcal{A}^{\beta}_0$, the difference between $\hat{A}_1(t)$ and $\hat{A}_2(t)$ at
	point $t_0$ is given by
	\begin{equation}
	\|\hat{A}_1(t_0)-\hat{A}_2(t_0)\|_2^2 \geq \frac{ \gamma^2L^2c_0^{2\beta}f^2(0)}{32}m^2\Big(\frac{mr}{n}\Big)^{\frac{2\beta}{2\beta+1}}. \label{ProofThm_7_2}
	\end{equation}
	On the other hand, we consider the joint distributions $P^A_{\tau,X,Y}$ such that $\tau\sim U[0,1]$, $X\sim \Pi_0$ where $\Pi_0$ denotes the uniform distribution on $\mathcal{X}$, $\tau$ and $X$ 
	are independent,
	and 
	$$
	\mathbb{P}_A(Y|\tau,X) =
	\begin{cases}
	\frac{1}{2}+\frac{\langle A(\tau), X\rangle}{4a}, & Y = a, \\
	\frac{1}{2}-\frac{\langle A(\tau), X\rangle}{4a}, & Y = -a .
	\end{cases}
	$$
	One can easily check that as long as $A(\tau)\in\mathcal{A}^{\beta}_0$, such $P^A_{\tau,X,Y}$ belongs to the distribution class $\mathcal{P}(r,a)$. We denote the corresponding $n-$product 
	probability measure
	by $\mathbb{P}_{A}$. 
	Then for any $A(\tau)\in\mathcal{A}^{\beta}_0$, the Kullback-Leibler Divergence between $\mathbb{P}_0$ and $\mathbb{P}_A$ is
	$$
	K(\mathbb{P}_0,\mathbb{P}_A) = n\mathbb{E}\Big( p_0(\tau,X)\log \frac{p_0(\tau,X)}{p_A(\tau,X)} + (1-p_0(\tau,X))\log \frac{1-p_0(\tau,X)}{1-p_A(\tau,X)}    \Big),
	$$
	where $p_A(\tau,X) = 1/2 + \langle A(\tau), X\rangle/4a$.
	Note that $\mathbb{P}_A(Y=a|\tau,X)\in [1/4,3/4]$ is guaranteed provided that $|\langle A(t), X\rangle| \leq a$. Thus by the inequality $-\log(1+u) \leq -u+u^2/2,~\forall u>-1$,
	and the fact that $\mathbb{P}_A(Y=a|\tau,X)\in [1/4,3/4]$, we have
	$$
	K(\mathbb{P}_0,\mathbb{P}_A)  \leq n\mathbb{E}2(p_0(\tau,X)-p_A(\tau,X))^2 \leq \frac{n}{8a^2}\mathbb{E}\langle A(\tau), X \rangle^2.
	$$
	Recall that $A(\tau) = Af_n(\tau) \in \mathcal{A}^{\beta}_0$, by $\tau \sim U[0,1]$ and $X\sim \Pi_0$, we have 
	\begin{equation}
	K(\mathbb{P}_0,\mathbb{P}_A)  \leq \frac{n}{8a^2} \frac{1}{m^2}L^2 \|f\|_2^2h_n^{2\beta+1} m^2\gamma ^2 \leq \frac{L^2 \|f\|_2^2 c_0^{2\beta+1}\gamma^2}{8a^2}mr. \label{ProofThm_7_3}
	\end{equation}
	Therefore, provided the fact that ${\rm Card}(\mathcal{A}^0)\geq 2^{mr/32} + 1$, together with (\ref{ProofThm_7_3}), we have
	\begin{equation}
	\frac{1}{{\rm Card}(\mathcal{A}^{\beta}_0)-1}\sum\limits_{A\in\mathcal{A}^{\beta}_0}K(\mathbb{P}_0, \mathbb{P}_A) \leq \alpha\log({\rm Card}(\mathcal{A}^{\beta}_0)-1) \label{ProofThm_7_4}
	\end{equation}
	is satisfied for any $\alpha>0$ if $\gamma$ is chosen as a sufficiently small constant. In view of (\ref{ProofThm_7_2}) and (\ref{ProofThm_7_4}), the lower bound (\ref{Thm_bound_7}) follows from Theorem 2.5 in \cite{tsybakov2009introduction}.
\end{proof}

\subsection{Proof of Theorem \ref{Thm_8}}
\label{ProofThm_8}
\begin{proof}
	Without loss of generality, we assume that both $m$ and $r$ are even numbers.
	Take a real number $c_1>0$, define 
	$$
	M = \ceil[\Big]{c_1\Big(\frac{n}{mr }\Big)^{\frac{1}{2\beta+1}}},~~~h_n = \frac{1}{2M},~~~t_j = \frac{2j-1}{2M},
	$$
	and
	$$
	\phi_{j}(t) = Lh_n^{\beta}f\Big(\frac{t-t_j}{h_n}\Big),~~~j=1,...M,~~~t\in[0,1],
	$$
	where $f$ is defined the same as in (\ref{ProofThm_7_5}). 
	Meanwhile, we consider the set of all binary sequences of length $M$: $\Omega = \Big\{ \omega = (\omega_1,...,\omega_M), \omega_i\in\{0,1\}\Big\} = \{0,1\}^M.$
	By Varshamov-Gilbert bound, there exists a subset $\Omega_0 = \{\omega^0,...,\omega^N\}$ of $\Omega$ such that $\omega^0 = (0,...,0)\in\Omega_0$, and
	$
	d(\omega^j,\omega^k) \geq \frac{M}{8},~\forall~0\leq j<k \leq N,
	$
	and 
	$
	N \geq 2^{\frac{M}{8}},
	$
	where $d(\cdot,\cdot)$ denotes the Hamming distance of two binary sequences.  
	Then we define a collection of functions based on $\Omega_0$:
	$
	\mathcal{E} = \Big\{ f_{\omega} (t) = \sum\limits_{j=1}^M \omega_j\phi_{j}(t) : \omega \in \Omega_0\Big\}.
	$
	From the result of Varshamov-Gilbert bound, we know that 
	$
	S:={\rm Card}(\mathcal{E}) = {\rm Card}(\Omega_0) \geq 2^{\frac{M}{8}}+1.
	$
	It is also easy to check that for all $f_{\omega},~f_{\omega'} \in\mathcal{E} $, 
	\begin{equation}
	\begin{aligned}
	\int_0^1 (f_{\omega}(t)-f_{\omega'}(t))^2 dt &= \sum\limits_{j=1}^M(\omega_j-\omega'_j)^2 \int_{\Delta_j} \phi_j^2(t)dt\\
	& = L^2 h_n^{2\beta+1} \|f\|_2^2\sum\limits_{j=1}^M(\omega_j-\omega'_j)^2\\
	&\geq L^2 h_n^{2\beta}\|f\|_2^2/16,
	\end{aligned}  \label{ProofThm_8_1}
	\end{equation}
	where $\Delta_j = [(j-1)/M,j/M]$.
	
	In what follows, we combine two fundamental results in coding theory: one is Varshamov-Gilbert bound (\cite{gilbert1952comparison,varshamov1957estimate}) in its general form of a q-ary code, the other is the volume estimate of Hamming balls.
	Let $A_q(n,d)$ denote the largest size of a $q$-ary code of block length $n$ with minimal Hamming distance $d$. 
	
	\begin{proposition}
		\label{Lemma_4}
		The maximal size of a $q-ary$ code of block length $n$ with minimal Hamming distance $d=pn$, satisfies
		\begin{equation}
		A_q(n,d+1) \geq q^{n(1-h_q(p))}, \label{vg}
		\end{equation}
		where $p\in[0,1-1/q]$, $h_q(p) = p\log_q(q-1)-p\log_qp-(1-p)\log_q(1-p)$ is the $q-ary$ entropy function.
	\end{proposition}
	
	We now have all the elements needed in hand to construct our hypotheses set. 
	Denote
	$
	\Omega_1 = \{\omega^1,...,\omega^N\},
	$
	which is a subset of $\Omega_0$ without $\omega^0$. We then consider a subset $\mathcal{E}_1$ of $\mathcal{E}$ which is given by
	$
	\mathcal{E}_1 := \Big\{ f_{\omega} (t) = \sum\limits_{j=1}^M \omega_j\phi_{j}(t) : \omega \in \Omega_1\Big\}.
	$
	Clearly, 
	$
	S_1 := {\rm Card}(\mathcal{E}_1) \geq 2^{M/8}.
	$
	Then we define a new collection of matrix valued functions as 
	$$
	\mathcal{C}=\big\{\tilde{A}=(a_{ij})\in \mathbb{C}^{\frac{m}{2}\times \frac{r}{2}}: a_{ij}\in\{\delta f_{\omega}:\omega \in \Omega_1 \big\},~\delta\in\CC, \forall 1\leq i\leq m/2, 1\leq j \leq r/2 \}.
	$$
	Obviously, the collection $\mathcal{C}$ is a $S_1$-ary code of block length $mr/4$. Thus, we can apply the result of Proposition \ref{Lemma_4}. It is easy to check that for $p=1/4$, and $q\geq 4$
	\begin{equation}
	\begin{aligned}
	1-h_q(p) &= 1 -p\log_q\frac{q-1}{p}+(1-p)\log_q(1-p) \geq \frac{1}{4}.
	\label{entropy}
	\end{aligned}
	\end{equation}
	In our case, $q= S_1\geq 2^{M/8}$ and $n=mr/4$. If we take $p=1/4$, we know that 
	\begin{equation}
	A_{S_1}(mr/4, mr/16)\geq A_{S_1}(mr/4,mr/16+1) \geq S_1^{mr/16}. \label{exist}
	\end{equation}
	In other words, (\ref{exist}) guarantees that there exists a subset $\mathcal{H}^0 \subset \mathcal{C}$ with ${\rm Card}(\mathcal{H}^0) \geq 2^{Mmr/128}$ such that for any $A_1,A_2\in\mathcal{H}^0$, the Hamming distance 
	between $A_1$ and $A_2$ is at least $mr/16$. Now we define the building blocks of our hypotheses set 
	$$
	\mathcal{H} :=  \mathcal{H}^0 \cup  \Big\{O_{\frac{m}{2}\times \frac{r}{2}}\Big\},
	$$
	where $O_{\frac{m}{2}\times \frac{r}{2}}$ is the $\frac{m}{2}\times \frac{r}{2}$ zero matrix. Obviously, $\mathcal{H}$ has size ${\rm Card}(\mathcal{H}) \geq 2^{M mr/64}+1$, and for any $A_1(t), A_2(t)\in\mathcal{H}$, the minimum Hamming distance is still greater than $mr/16$. 
	We consider the set of matrix valued functions
	$$
	\mathcal{B}(\mathcal{H})
	= \Bigg\{
	\begin{bmatrix}
	\tilde{A} & \tilde{A} & \dots & \tilde{A} & O \\
	\end{bmatrix}
	: \tilde{A}\in \mathcal{H}
	\Bigg\},
	$$
	where $O$ denotes the $ m/2\times (m/2-r\lfloor m/r\rfloor/2)$ zero matrix. Finally, our hypotheses set of matrix valued functions $\mathcal{H}_m$ is defined as
	$$
	\mathcal{H}_m
	= \Bigg\{
	\begin{bmatrix}
	\tilde{O} & \hat{A} \\
	\hat{A}^{*} & \tilde{O}
	\end{bmatrix}
	\in \mathbb{C}^{m \times m}: \hat{A}\in \mathcal{B}(\mathcal{H})
	\Bigg\}.
	$$
	By the definition of $\mathcal{H}_m$ and similar to the arguments in proof of Theorem \ref{Thm_7}, it is easy to check that $\mathcal{H}_m\subset \mathcal{A}(r,a) $, and also
	\begin{equation}
	{\rm Card}(\mathcal{H}_m) \geq 2^{Mmr/64} +1 .\label{ProofThm_8_3}
	\end{equation}
	
	Now we consider any two different hypotheses $A_j(t), A_k(t)\in \mathcal{H}_m$. 
	\begin{equation}
	\int_0^1 \|A_j(t)- A_k(t)\|_2^2 dt  \geq  \gamma^2\frac{mr}{16} 2 \Big\lfloor \frac{m}{r} \Big\rfloor \int^1_0 (f_{\omega}(t)-f_{\omega'}(t))^2dt,
	\end{equation}
	where $\omega \neq \omega'$. Based on (\ref{ProofThm_8_1}), we have
	\begin{equation}
	\frac{1}{m^2}\int_0^1 \|A_j(t)- A_k(t)\|_2^2 dt  \geq \frac{\gamma^2L^2h_n^{2\beta}\|f\|_2^2}{256} \geq c_* \Big(\frac{mr}{n}\Big)^{\frac{2\beta}{2\beta+1}}. \label{ProofThm_8_5}
	\end{equation}
	where $c_*$ is a constant depending on $\|f\|_2$, $L$, $c_1$ and $\gamma$. 
	
	On the other hand, we repeat the same analysis on the Kullback-Leibler divergence $K(\mathbb{P}_0, \mathbb{P}_A)$ as in the proof of Theorem \ref{Thm_7}.
	One can get
	\begin{equation}
	K(\mathbb{P}_0, \mathbb{P}_A) \leq \frac{n}{8a^2}\mathbb{E}\langle A(\tau) ,X\rangle^2 \leq \frac{n}{8a^2}\gamma^2  \sum\limits_{j=1}^M \int^1_0\phi_j^2(\tau)d\tau   \leq  \frac{\gamma^2c_1^{2\beta+1} L^2 Mmr \|f\|^2_2}{8a^2}, \label{ProofThm_8_2}
	\end{equation}
	where $A(\tau) \in \mathcal{H}_m$.
	Combine  (\ref{ProofThm_8_3}) and (\ref{ProofThm_8_2}) we know that
	\begin{equation}
	\frac{1}{{\rm Card}(\mathcal{H}_m)-1}\sum\limits_{A(t)\in\mathcal{H}_m}K(\mathbb{P}_0, \mathbb{P}_A) \leq \alpha\log({\rm Card}(\mathcal{H}_m)-1) \label{ProofThm_8_4}
	\end{equation}
	is satisfied for any $\alpha>0$ if $\gamma$ is chosen as a sufficiently small constant. In view of (\ref{ProofThm_8_5}) and (\ref{ProofThm_8_4}), the lower bound follows from Theorem 2.5 in \cite{tsybakov2009introduction}.	
\end{proof}

\subsection{Proof of Theorem \ref{Thm_9}}
\label{ProofThm_9}
\begin{proof}
	Without loss of generality, assume that $m$ is an even number.
	For some constant $\gamma >0$, denote
	$
	\mathcal{V}=\Big\{v\in \mathbb{C}^{\frac{m}{2}}: a_{i}\in\{0,\gamma \},\forall~1\leq i\leq m/2 \Big\}.
	$
	Due to the Varshamov-Gilbert bound (see
	Lemma 2.9 in \cite{tsybakov2009introduction}), there exists a subset $\mathcal{V}^0 \subset \mathcal{V}$ with cardinality ${\rm Card}(\mathcal{V}^0)\geq 2^{m/16} + 1$ containing the zero vector
	$\mathbf{0} \in \mathbb{C}^{\frac{m}{2}}$, and such that for any two distinct elements $v_1$ and $v_2$ of $\mathcal{V}^0$,
	\begin{equation}
	\|v_1-v_2\|_2^2 \geq \frac{m}{16} \gamma^2. 
	\end{equation}
	Consider the set of matrices
	$$
	\mathcal{B}(\mathcal{V})
	= \Big\{
	\begin{bmatrix}
	v & v & \dots & v &  \\
	\end{bmatrix}
	\in \mathbb{C}^{\frac{m}{2}\times \frac{m}{2}}: v \in \mathcal{V}_0
	\Big\}.
	$$
	Clearly, $\mathcal{B}(\mathcal{V})$ is a collection of rank one matrices.
	Then we construct another matrix set $\mathcal{V}_m$,
	$$
	\mathcal{V}_m
	= \Bigg\{
	\begin{bmatrix}
	\tilde{O} & V \\
	V^* & \tilde{O}
	\end{bmatrix}
	\in \mathbb{C}^{m \times m}: V\in \mathcal{B}(\mathcal{V})
	\Bigg\}
	$$
	where $\tilde{O}$ is the $m/2\times m/2$ zero matrix. Apparently, $\mathcal{V}_m \subset \mathbb{H}_m$.
	
	On the other hand, we define the grid on $[0,1]$
	$$
	M = \ceil[\Big]{c_2\Big(\frac{n}{m+\log n }\Big)^{\frac{1}{2\beta+1}}},~~~h_n = \frac{1}{2M},~~~t_j = \frac{2j-1}{2M},
	$$
	and
	$$
	\phi_{j}(t) = Lh_n^{\beta}f\Big(\frac{t-t_j}{h_n}\Big),~~~j=1,...M,~~~t\in[0,1]
	$$
	where $f$ is defined the same as in (\ref{ProofThm_7_5}), and $c_2$ is some constant. Denote
	$
	\Phi:=\Big\{\phi_j:j=1,...M\Big\}.
	$
	We consider the following set of hypotheses:
	$
	\mathcal{A}^{\beta}_{\mathcal{B}}:=\{\hat{A}(t)=V \phi_j(t): V\in \mathcal{V}_m,~\phi_j\in \Phi \}.
	$
	One can immediately get that the size of $\mathcal{A}^{\beta}_{\mathcal{B}}$ satisfies 
	\begin{equation}
	{\rm Card}(\mathcal{A}^{\beta}_{\mathcal{B}})\geq (2^{m/16}+1)M. \label{ProofThm_9_3}
	\end{equation}
	By construction, the following claims are obvious: any element $\hat{A}(t)$
	of $\mathcal{A}^{\beta}_{\mathcal{B}}$ has rank at most $2$; the
	entries of $\hat{A}(t) \in \mathcal{A}^{\beta}_{\mathcal{B}}$ are uniformly bounded for some sufficiently small $\gamma$, and $\hat{A}_{ij}(t)\in \Sigma(\beta, L)$. Thus $\mathcal{A}^{\beta}_{\mathcal{B}}\subset \mathcal{A}(a)$. 
	
	Now we consider the distance between two distinct elements $A(t)$ and $A'(t)$ of $\mathcal{A}^{\beta}_{\mathcal{B}}$. An immediate observation is that 
	$$
	\sup\limits_{t\in[0,1]} \|A(t)-A'(t)\|^2 \geq \frac{1}{4}\sup\limits_{t\in[0,1]} \|A(t)-A'(t)\|_2^2,
	$$
	due to the fact that $\forall t\in(0,1)$, ${\rm rank}(A(t)-A'(t)) \leq 4$. Then we turn to get lower bound on $\sup\limits_{t\in(0,1)} \|A(t)-A'(t)\|_2^2$.
	Recall that by construction of $\mathcal{A}^{\beta}_{\mathcal{B}}$, we have for any $A \neq A'$,
	$
	A(t) = A_1\phi_j(t),~A'(t) = A_2\phi_k(t),
	$
	where $A_1, A_2\in \mathcal{V}_m$. 
	There are three cases need to be considered: 1). $A_1 \neq A_2$ and $j =k$; 2). $A_1 = A_2 \neq 0$ and $j \neq k$; 3). $A_1 \neq A_2$ and $j \neq k$.
	
	For case 1, 
	$$
	\begin{aligned}
	\sup\limits_{t\in [0,1]} \|A(t)- A'(t)\|_2^2  &= \|A_1 - A_2\|_2 ^2 \|\phi_j\|_{\infty}^2 \\
	&\geq  \frac{m^2}{16}  \gamma^2 L^2 h_n^{2\beta}\|f\|_{\infty}^2 \geq c_1^* m^2 \Big(\frac{m+\log n}{n}\Big)^{\frac{2\beta}{2\beta+1}},
	\end{aligned}
	$$
	where $c_1^*$ is a constant depending on $\|f\|_{\infty}^2$, $\beta$, $L$ and $\gamma$.
	
	For case 2,
	$$
	\begin{aligned}
	\sup\limits_{t\in [0,1]} \|A(t)- A'(t)\|_2^2  &= \|A_1 \|_2 ^2 \|\phi_j-\phi_k\|_{\infty}^2 \\
	&\geq  \frac{m^2}{16}  \gamma^2 L^2 h_n^{2\beta}\|f\|_{\infty}^2 \geq c_2^* m^2  \Big(\frac{m+\log n}{n}\Big)^{\frac{2\beta}{2\beta+1}},
	\end{aligned}
	$$
	where $c_2^*$ is a constant depending on $\|f\|_{\infty}^2$, $\beta$, $L$ and $\gamma$.
	
	For case 3,
	$$
	\begin{aligned}
	\sup\limits_{t\in [0,1]} \|A(t)- A'(t)\|_2^2  &\geq  (\|A_1\|_2 ^2 \|\phi_j\|_{\infty}^2 \vee \|A_2\|_2 ^2 \|\phi_k\|_{\infty}^2) \\
	& \geq  \frac{m^2}{16}  \gamma^2 L^2 h_n^{2\beta}\|f\|_{\infty}^2 \geq c_3^* m^2 \Big(\frac{m+\log n}{n}\Big)^{\frac{2\beta}{2\beta+1}},
	\end{aligned}
	$$
	where $c_3^*$ is a constant depending on $\|f\|_{\infty}^2$, $\beta$, $L$ and $\gamma$.
	
	Therefore, by the analysis above we conclude that for any two distinct elements $A(t)$ and $A'(t)$ of $\mathcal{A}^{\beta}_{\mathcal{B}}$,
	\begin{equation}
	\sup\limits_{t\in[0,1]} \|A(t)-A'(t)\|^2 \geq \frac{1}{4}\sup\limits_{t\in[0,1]} \|A(t)-A'(t)\|_2^2 \geq  c_* m^2 \Big(\frac{m+\log n}{n}\Big)^{\frac{2\beta}{2\beta+1}}, \label{ProofThm_9_5}
	\end{equation}
	where $c_*$ is a constant depending on $\|f\|_{\infty}^2$, $L$, $\gamma$ and $\beta$.
	
	Meanwhile, we repeat the same analysis on the Kullback-Leibler divergence $K(\mathbb{P}_0, \mathbb{P}_A)$ as in the proof of Theorem \ref{Thm_7}.
	One can get that for any $A\in \mathcal{A}^{\beta}_{\mathcal{B}}$, the Kullback-Leibler divergence $K(\mathbb{P}_0, \mathbb{P}_A)$
	between $\mathbb{P}_0$ and $\mathbb{P}_A$ satisfies
	
	\begin{equation}
	K(\mathbb{P}_0, \mathbb{P}_A) \leq \frac{n}{8a^2}\mathbb{E}|\langle A(\tau) ,X\rangle|^2 \leq \frac{n}{8a^2}\gamma^2   \int^1_0\phi_j^2(\tau)d\tau   \leq  \frac{\gamma^2c_2^{2\beta+1} L^2 (m+\log n) \|f\|^2_2}{8a^2}. \label{ProofThm_9_2}
	\end{equation}
	
	Combine  (\ref{ProofThm_9_3}) and (\ref{ProofThm_9_2}) we know that
	\begin{equation}
	\frac{1}{{\rm Card}(\mathcal{A}^{\beta}_{\mathcal{B}})-1}\sum\limits_{A\in \mathcal{A}^{\beta}_{\mathcal{B}}}K(\mathbb{P}_0, \mathbb{P}_A) \leq \alpha\log({\rm Card}(\mathcal{A}^{\beta}_{\mathcal{B}})-1) \label{ProofThm_9_4}
	\end{equation}
	is satisfied for any $\alpha>0$ if $\gamma$ is chosen as a sufficiently small constant. In view of (\ref{ProofThm_9_5}) and (\ref{ProofThm_9_4}), the lower bound follows from Theorem 2.5 in 
	\cite{tsybakov2009introduction}.
\end{proof}

\subsection{Proof of Theorem \ref{Thm_10}}
\label{Proof:Thm_10}
\begin{proof}
	For any $\hat{A}^k$, denote the difference in empirical loss between $\hat{A}^k$ and $A$ by 
	\begin{align*}
	r_n(\hat{A}^k,A) :&= \frac{1}{n} \sum\limits_{j=1}^{n}(Y_j-\langle \hat{A}^k(\tau_j), X_j\rangle)^2 - \frac{1}{n} \sum\limits_{j=1}^{n}(Y_j-\langle A(\tau_j), X_j\rangle)^2  
	= -\frac{1}{n} \sum\limits_{j=1}^{n} U_j,
	\end{align*}
	where $U_j = (Y_j-\langle A(\tau_j), X_j\rangle)^2 - (Y_j-\langle \hat{A}^k(\tau_j), X_j\rangle)^2$. 
	It is easy to check that 
	\begin{equation}
	\label{U_j}
	U_j = 2(Y_j - \langle A(\tau_j) ,X_j\rangle)\langle \hat{A}^k(\tau_j)- A(\tau_j), X_j\rangle -  \langle \hat{A}^k(\tau_j)- A(\tau_j), X_j\rangle^2.
	\end{equation}
	We denote
	$
	r(\hat{A}^k,A) :=  \EE \langle \hat{A}^k(\tau) - A(\tau), X \rangle^2.
	$
	The following concentration inequality developed by \cite{craig1933tchebychef} to prove Bernstein's inequality is key to our proof. 
	\begin{lemma}
		\label{Bernstein_ineq}
		Let $U_j$, $j=1,...,n$ be independent bounded random variables satisfying $|U_j -\EE U_j| \leq M$ with $h=M/3$. 
		Set $\bar{U} = n^{-1} \sum^{n}_{j=1} U_j$. Then for all $t>0$
		$$
		\PP\Big\{ \bar{U} - \EE\bar{U} \geq \frac{t}{n\varepsilon} + \frac{n\varepsilon {\rm var} (\bar{U})}{2(1-c)}\Big\} \leq e^{-t},
		$$
		with $0<\varepsilon h \leq c<1$.
	\end{lemma}
	Firstly, we bound the variance of $U_j$. Under the assumption that $|Y|$ and $|\langle A(\tau), X\rangle|$ are bounded by a constant $a$, one can easily check that $h = 8a^2/3$.
	Given $\EE (Y_j | \tau_j , X_j) = \langle A(\tau_j) , X_j \rangle$, we know that the covariance between the two terms on the right hand side of (\ref{U_j}) is zero. Conditionally on $(\tau,X)$, the second order moment of the first term satisfies
	$$
	4\EE \sigma^2_{Y|\tau,X} \langle \hat{A}^k(\tau_j)- A(\tau_j), X_j\rangle^2 \leq 4a^2 r(\hat{A}^k,A) .
	$$
	To see why, one can consider 
	the random variable $\tilde{Y}$ with the distribution $\PP\{ \tilde{Y} = a\} = \PP\{ \tilde{Y} = -a\} = 1/2$. The variance of $Y$ is always bounded by the variance of $\tilde{Y}$ which is $a^2$ 
	under the assumption that $|Y_j|$ and $|\langle \hat{A}^k(\tau_j) , X_j \rangle|$ are bounded by a constant $a>0$. Similarly, we can get that the variance of the second term conditioned on 
	$(\tau,X)$ is also bounded by 
	$4a^2 \EE \langle \hat{A}^k(\tau_j)- A(\tau_j), X_j\rangle^2 $. As a result,
	$
	n{\rm var}(\bar{U}) \leq 8a^2 r(\hat{A}^k,A).
	$
	By the result of Lemma \ref{Bernstein_ineq}, we have for any $\hat{A}^k$ with probability at least $1-e^{-t}$
	$$
	r(\hat{A}^k,A)- r_n(\hat{A}^k,A) < \frac{t}{n\varepsilon} + \frac{4a^2 \varepsilon r(\hat{A}^k,A)}{1-c}.
	$$
	Set $t = \varepsilon \pi_k+\log 1/\delta$, we get with probability at least $1-\delta/e^{\varepsilon\pi_k}$
	$$
	(1 - \alpha )  r(\hat{A}^k,A)  <  r_n(\hat{A}^k,A) + \frac{\pi_k}{n} + \frac{4a^2}{(1-c)\alpha}\Big(\frac{\log1/\delta}{n}\Big).
	$$
	where $\alpha  = 4a^2\varepsilon/(1-c) < 1$. Denote
	$$
	\tilde{k}^* = \arg\min\limits_{k}  \Big\{ r(\hat{A}^k,A)   + \frac{\pi_k}{n} \Big\}.
	$$
	By the definition of $\hat{A}^*$, we have with probability at least $1-\delta/e^{\varepsilon \hat{\pi}^*}$
	\begin{equation}
	\label{adp_eq_1}
	(1 - \alpha )  r(\hat{A}^*,A) < r_n(\hat{A}^{\tilde{k}^*},A) +  \frac{\pi_{\tilde{k}^*}}{n} + \frac{4a^2}{(1-c)\alpha} \Big(\frac{\log1/\delta}{n}\Big).
	\end{equation}
	where $\hat{\pi}^*$ is the penalty terms associated with $\hat{A}^*$.
	
	Now we apply the result of Lemma \ref{Bernstein_ineq} one more time and set $t = \log 1/\delta$, we get with probability at least $1-\delta$
	\begin{equation}
	\label{adp_eq_2}
	r_n(\hat{A}^{\tilde{k}^*},A) \leq (1+\alpha)r(\hat{A}^{\tilde{k}^*},A) +  \frac{4a^2}{(1-c)\alpha} \frac{\log 1/\delta}{n}.
	\end{equation}
	Apply the union bound of (\ref{adp_eq_1}) and (\ref{adp_eq_2}), we get with probability at least $1-\delta(1+e^{-\varepsilon\hat{\pi}^*})$
	$$
	r(\hat{A}^*,A) \leq \frac{(1+\alpha)}{(1-\alpha)} \Big( r(\hat{A}^{\tilde{k}^*},A)  + \frac{\pi_{\tilde{k}^*}}{n} \Big)+ \frac{4a^2}{(1-c)\alpha(1-\alpha)}\frac{\log 1/\delta}{n}.
	$$
	By taking $\varepsilon = 3/32a^2$ and $c = \varepsilon h$, 
	$$
	r(\hat{A}^*,A) \leq 3 \Big( r(\hat{A}^{\tilde{k}^*},A)  + \frac{\pi_{\tilde{k}^*}}{n} \Big)+ \frac{64a^2}{3}\frac{\log 1/\delta}{n}.
	$$
	By taking $\delta = 1/ n^{mr}$ and adjusting the constant, we have with probability at least $1-1/n^{mr}$
	$$
	\frac{1}{m^2}  \int_0^1\big\| \hat{A}^*(t) - A(t)\big\|^2_2dt \leq 3\min\limits_k\Big\{\frac{1}{m^2} \int_0^1\big\| \hat{A}^k(t) - A(t)\big\|^2_2dt + \frac{\pi_k}{n} \Big\} +  C(a)\frac{mr\log n}{n}
	$$
	where $C(a)$ is a constant depending on $a$.
\end{proof}

\section*{Acknowledgements}
The author would like to thank Dr. Vladimir Koltchinskii for all the guidance, discussion and inspiration along the course of writing this paper. The author also would like to thank Dr. Dong Xia for several insightful 
comments and some meaningful discussions on this paper.
The author would like to thank NSF for its generous support through Grants DMS-1509739 and CCF-1523768.

\bibliographystyle{plainnat}
\bibliography{refer}

\section{Appendix: Proof of Lemma 1}
\label{proof:Thm_1}
The proof of Lemma \ref{Thm_1} follows from a similar approach introduced by \cite{koltchinskii2013sharp}.
\begin{proof}
	For any $S\in\mathbb{H}_m$ of ${\rm rank}$ $r$, $S=\sum_{j=1}^r \lambda_i(e_j\otimes e_j)$, where $\lambda_j$ are non-zero eigenvalues of $S$ (repeated with their multiplicities) and $e_j\in \mathbb{C}^m$ are the corresponding orthonormal eigenvectors. Denote ${\rm sign}(S):= \sum\limits_{j=1}^r {\rm sign}(\lambda_i)(e_j\otimes e_j).$
	Let $\mathcal{P}_L$, $\mathcal{P}^{\perp}_L$ be the following orthogonal projectors in the space ($\mathbb{H}_m, \langle\cdot,\cdot\rangle$):
	$$
	\mathcal{P}_L(A):=A-P_{L^\perp}AP_{L^\perp},~\mathcal{P}_L^{\perp}(A):=P_{L^\perp}AP_{L^\perp},~\forall A\in \mathbb{H}_m
	$$
	where $P_L$ denotes the orthogonal projector on the linear span of $\{e_1,...,e_r\}$, and $P_{L^{\perp}}$ is its orthogonal complement.
	Clearly, this formulation provides a decomposition of a matrix A into a "low rank part" $\mathcal{P}_L(A)$ and a "high rank part" $\mathcal{P}_L^{\perp}(A)$ if ${\rm rank}(S)=r$ is small. 
	Given $b>0$, define the following cone in the space $\mathbb{H}_m$:
	$$
	\mathcal{K}(\mathbb{D};L;b):= \{A\in \DD: \|\mathcal{P}_L^{\perp}A\|_1\leq b\|\mathcal{P}_L(A)\|_1\}
	$$
	which consists of matrices with a "dominant" low rank part if $S$ is low rank. 
	
	Firstly, we can rewrite (\ref{Estimator}) as 
	\begin{equation}
	\label{Reform:Estimator}
	\hat{S}^h =\arg\min\limits_{S\in \mathbb{D}} \frac{1}{n}\sum\limits_{j=1}^{n} \Big(\tilde{Y}_j-\Big\langle  S , \tilde{X}_j\Big\rangle\Big)^2 +\varepsilon \|S\|_1. 
	\end{equation}
	where $\tilde{X}_j = {\rm Diag}\Big[\sqrt{\frac{1}{h}K\Big(\frac{\tau_j-t_0}{h}\Big)}p_0\Big(\frac{\tau_j-t_0}{h}\Big) X_j,...,\sqrt{\frac{1}{h}K\Big(\frac{\tau_j-t_0}{h}\Big)}p_{\ell}\Big(\frac{\tau_j-t_0}{h}\Big)X_j\Big]$, and
	$\tilde{Y}_j = \sqrt{\frac{1}{h}K\Big(\frac{\tau_j-t_0}{h}\Big)}Y_j$.
	
	Denote the loss function as
	$$
	\mathcal{L}\big(\tilde{Y};\langle S(\tau),\tilde{X}\rangle \big) :=  \Big(\tilde{Y}_j-\Big\langle  S , \tilde{X}_j\Big\rangle\Big)^2,
	$$
	and the risk 
	$$
	P\mathcal{L}\big(\tilde{Y};\langle S(\tau),\tilde{X}\rangle \big)  := \EE\mathcal{L}\big(\tilde{Y};\langle S(\tau),\tilde{X}\rangle \big)= \sigma^2 + \EE\frac{1}{h}K\Big( \frac{\tau-t_0}{h}\Big)\big(Y-\langle S(\tau),X\rangle  )^2
	$$
	
	Since $\hat{S}^h$ is a solution of the convex optimization problem (\ref{Reform:Estimator}), there exists a $\hat{V}\in\partial \big\|\hat{S}^h\big\|_1$, such that for $\forall S\in \mathbb{D}$ (see \cite{aubin2006applied} Chap. 2)
	$$
	\frac{2}{n}\sum\limits_{j=1}^n \Big(\Big\langle \hat{S}^h, \tilde{X}_j \Big\rangle-\tilde{Y}_j\Big) \langle \hat{S}^h-S ,\tilde{X_j}\rangle+ \varepsilon \langle \hat{V}, \hat{S}^h -S \rangle \leq 0.
	$$
	This implies that, for all $S\in \mathbb{D}$,
	\begin{equation}
	\begin{aligned}
	&\mathbb{E}\mathcal{L}'(\tilde{Y};\langle \hat{S}^h, \tilde{X}\rangle)\langle\hat{S}^h-S, \tilde{X}\rangle +  \varepsilon \langle \hat{V}, \hat{S}^h-S \rangle \\
	&\leq \mathbb{E}\mathcal{L}'(\tilde{Y};\langle \hat{S}^h, \tilde{X}\rangle)\langle\hat{S}^h-S, \tilde{X}\rangle- \frac{2}{n}\sum\limits_{j=1}^n (\langle \hat{S}^h , \tilde{X}_j\rangle-\tilde{Y}_j) \langle\hat{S}^h-S , \tilde{X}_j\rangle. \label{ProofThm_1_1}
	\end{aligned}
	\end{equation}
	where $\mathcal{L}'$ denotes the partial derivative of $\mathcal{L}(y;u)$ with respect to $u$.
	One can easily check that for $\forall S\in \mathbb{D}$,
	\begin{equation}
	\begin{aligned}
	\mathbb{E}\mathcal{L}'(\tilde{Y};\langle \hat{S}^h, \tilde{X}\rangle)\langle\hat{S}^h-S, \tilde{X}\rangle \geq  \mathbb{E}(\mathcal{L}(\tilde{Y};\langle \hat{S}^h, \tilde{X}\rangle)- \mathcal{L}(\tilde{Y};\langle S, \tilde{X}\rangle))
	+ \|\hat{S}^h-S\|_{L_2(\tilde{\Pi})}^2. \label{ProofThm_1_2}
	\end{aligned}
	\end{equation}
	where $\tilde{\Pi}$ denotes the distribution of $\tilde{X}$.
	If $\mathbb{E}\mathcal{L}(\tilde{Y};\langle \hat{S}^h, \tilde{X}\rangle) \leq \mathbb{E}\mathcal{L}(\tilde{Y};\langle S, \tilde{X}\rangle)$ for $\forall S\in\mathbb{D}$, then the oracle inequality in Lemma \ref{Thm_1} holds trivially. 
	So we assume that $\mathbb{E}\mathcal{L}(\tilde{Y};\langle \hat{S}^h, \tilde{X}\rangle)> \mathbb{E}\mathcal{L}(\tilde{Y};\langle S, \tilde{X}\rangle)$ for some $S\in \mathbb{D}$. Thus, inequalities (\ref{ProofThm_1_1}) and (\ref{ProofThm_1_2}) imply that
	\begin{equation}
	\begin{aligned}
	&\mathbb{E}\mathcal{L}(\tilde{Y};\langle \hat{S}^h, \tilde{X}\rangle)+ \|\hat{S}^h-S\|_{L_2(\tilde{\Pi})} ^2 +  \varepsilon \langle \hat{V}, \hat{S}^h-S \rangle\\
	&\leq \mathbb{E}\mathcal{L}(\tilde{Y};\langle S, \tilde{X}\rangle) + \mathbb{E}\mathcal{L}'(\tilde{Y};\langle \hat{S}^h, \tilde{X}\rangle)\langle\hat{S}^h-S, \tilde{X}\rangle - \frac{2}{n}\sum\limits_{j=1}^n (\langle \hat{S}^h , \tilde{X}_j\rangle-\tilde{Y}_j) \langle\hat{S}^h-S , \tilde{X}_j\rangle. \label{ProofThm_1_3}
	\end{aligned}
	\end{equation}
	According to the well known representation of subdifferential of nuclear norm, see \cite{koltchinskii2011introduction} Sec. A.4, for any $V\in \partial \|S\|_1$, we have
	$$
	V :={\rm sign}(S)+\mathcal{P}^{\perp}_L(W),~W\in \mathbb{H}_m~,\|W\|\leq 1.
	$$
	By the duality between nuclear norm and operator norm
	$$
	\langle \mathcal{P}_L^{\perp}(W), \hat{S}^h-S\rangle = \langle \mathcal{P}_L^{\perp}(W), \hat{S}^h\rangle = \langle W, \mathcal{P}_L^{\perp}(\hat{S}^h)\rangle=\|\mathcal{P}_L^{\perp}(\hat{S}^h)\|_1. \label{4}
	$$
	Therefore, by the monotonicity of subdifferentials of convex function $\|\cdot\|_1$, for any $V :={\rm sign}(S)+\mathcal{P}^{\perp}_L(W)\in \partial\|S\|_1$, we have
	\begin{equation}
	\langle V,\hat{S}^h-S\rangle = \langle {\rm sign}(S), \hat{S}^h-S\rangle +\|\mathcal{P}_L^{\perp}(\hat{S}^h)\|_1 \leq \langle\hat{V},\hat{S}^h-S\rangle , \label{ProofThm_1_4}
	\end{equation}
	we can use (\ref{ProofThm_1_4}) to change the bound in (\ref{ProofThm_1_3}) to get
	\begin{equation}
	\begin{aligned}
	&\mathbb{E}\mathcal{L}(\tilde{Y};\langle \hat{S}^h, \tilde{X}\rangle) + \|S-\hat{S}^h\|_{L_2(\tilde{\Pi})}^2 +  \varepsilon \|\mathcal{P}_L^{\perp}(\hat{S}^h)\|_1\\
	&\leq \mathbb{E}\mathcal{L}(\tilde{Y};\langle S, \tilde{X}\rangle) + \varepsilon \langle {\rm sign}(S),S- \hat{S}^h\rangle + \mathbb{E}\mathcal{L}'(\tilde{Y};\langle \hat{S}^h, \tilde{X}\rangle) \langle\hat{S}^h-S, \tilde{X}\rangle\\
	&-\frac{2}{n}\sum\limits_{j=1}^n (\langle \hat{S}^h , \tilde{X}_j\rangle-\tilde{Y}_j) \langle\hat{S}^h-S , \tilde{X}_j\rangle. \label{ProofThm_1_5}
	\end{aligned}
	\end{equation}
	For the simplicity of representation, we use the following notation to denote the empirical process:
	\begin{equation}
	\begin{aligned}
	&(P-P_n)(\mathcal{L}'(\tilde{Y};\langle \hat{S}^h, \tilde{X}\rangle))\langle\hat{S}^h-S, \tilde{X}\rangle :=\\
	&\mathbb{E}\mathcal{L}'(\tilde{Y};\langle \hat{S}^h, \tilde{X}\rangle) \langle\hat{S}^h-S, \tilde{X}\rangle 
	- \frac{2}{n}\sum\limits_{j=1}^n  (\langle \hat{S}^h , \tilde{X}_j\rangle-\tilde{Y}_j) \langle\hat{S}^h-S , \tilde{X}_j\rangle. \label{ProofThm_1_6}
	\end{aligned}
	\end{equation}
	The following part of the proof is to derive an upper bound on the empirical process (\ref{ProofThm_1_6}). 
	Before we start with the derivation, let us present several vital ingredients that will be used in the following literature. For a given $S\in \mathbb{D}$ and for $\delta_1,\delta_2,\delta_3,\delta_4\geq 0$, denote
	$$
	\mathcal{A}(\delta_1,\delta_2):= \{A\in\mathbb{D}: A-S\in \mathcal{K}(\mathbb{D};L;b), \|A-S\|_{L_2(\tilde{\Pi})}\leq \delta_1, \|\mathcal{P}_L^{\perp}A\|_1\leq \delta_2 \},
	$$
	$$
	\tilde{\mathcal{A}}(\delta_1,\delta_2,\delta_3):= \{A\in\mathbb{D}: \|A-S\|_{L_2(\tilde{\Pi})}\leq \delta_1, \|\mathcal{P}_L^{\perp}A\|_1\leq \delta_2, \|\mathcal{P}_L(A-S)\|_1\leq \delta_3 \},
	$$
	$$
	\check{\mathcal{A}}(\delta_1,\delta_4):= \{A\in\mathbb{D}: \|A-S\|_{L_2(\tilde{\Pi})}\leq \delta_1, \|A-S\|_1\leq \delta_4  \},
	$$
	and
	$$
	\alpha_n(\delta_1,\delta_2):= \sup\{ |(P-P_n)(\mathcal{L}'(\tilde{Y};\langle A, \tilde{X}\rangle))\langle A-S, \tilde{X}\rangle|: A\in \mathcal{A}(\delta_1,\delta_2) \},
	$$
	$$
	\tilde{\alpha}_n(\delta_1,\delta_2,\delta_3):= \sup\{ |(P-P_n)(\mathcal{L}'(\tilde{Y};\langle A, \tilde{X}\rangle))\langle A-S, \tilde{X}\rangle|: A\in \tilde{\mathcal{A}}(\delta_1,\delta_2,\delta_3) \},
	$$
	$$
	\check{\alpha}_n(\delta_1,\delta_4):= \sup\{ |(P-P_n)(\mathcal{L}'(\tilde{Y};\langle A, \tilde{X}\rangle))\langle A-S, \tilde{X}\rangle|: A\in \check{\mathcal{A}}(\delta_1,\delta_4) \}.
	$$
	Given the definitions above, Lemma \ref{Lemma_1} below shows upper bounds on the three quantities $\alpha_n(\delta_1,\delta_2)$, $\tilde{\alpha}_n(\delta_1,\delta_2,\delta_3)$, $\check{\alpha}_n(\delta_1,\delta_4)$. The proof of Lemma \ref{Lemma_1} can be found in section \ref{Proof_Lemma_1}. 
	Denote 
	\begin{equation}
	\Xi :=n^{-1}\sum\limits_{j=1}^n\varepsilon_j  \tilde{X}_j \label{xi}
	\end{equation}
	where $\varepsilon_j$ are i.i.d. Rademacher random variables.
	\begin{lemma} 
		\label{Lemma_1}
		Suppose $0<\delta_k^-<\delta_k^+$, $k=1,2,3,4$. Let $\eta>0$ and
		$$
		\bar{\eta}:=\eta+\sum\limits_{k=1}^2\log([\log_2(\frac{\delta_k^+}{\delta_k^-})]+2)+\log3,
		$$
		$$
		\tilde{\eta}:=\eta+\sum\limits_{k=1}^3\log([\log_2(\frac{\delta_k^+}{\delta_k^-})]+2)+\log3,
		$$
		$$
		\check{\eta}:=\eta+\sum\limits_{k=1,k=4}\log([\log_2(\frac{\delta_k^+}{\delta_k^-})]+2)+\log3.
		$$
		Then with probability at least $1-e^{-\eta}$, for all $\delta_k\in [\delta_k^-, \delta_k^+]$, k=1,2,3
		\begin{equation}
		\alpha_n(\delta_1,\delta_2) \leq  \frac{C_1(\ell+1)R(T)\Phi a }{\sqrt{h}} \Big\{\mathbb{E}\|\Xi\|(\sqrt{{\rm rank}(S)} m\delta_1+\delta_2) +  \frac{2(\ell+1)R(T)\Phi a \bar{\eta} }{n\sqrt{h}}+  \delta_1 \sqrt{\frac{\bar{\eta}}{n}} \Big\}
		\end{equation}
		\begin{equation}
		\tilde{\alpha}_n(\delta_1,\delta_2,\delta_3) \leq  \frac{C_2(\ell+1)R(T)\Phi a }{\sqrt{h}}\Big\{ \mathbb{E}\|\Xi\|(\delta_2+\delta_3)+\frac{2(\ell+1)R(T)\Phi a\tilde{\eta}}{n\sqrt{h}}+  \delta_1\sqrt{\frac{\tilde{\eta}}{n}}\Big\}
		\end{equation} \label{410}
		\begin{equation}
		\check{\alpha}_n(\delta_1,\delta_4) \leq  \frac{C_3(\ell+1)R(T)\Phi a }{\sqrt{h}}\Big\{ \mathbb{E}\|\Xi\|\delta_4+\frac{2(\ell+1)R(T)\Phi a\check{\eta}}{n\sqrt{h}}+  \delta_1\sqrt{\frac{\check{\eta}}{n}}\Big\}
		\end{equation} \label{17}
		where $C_1$, $C_2$, and $C_3$ are numerical constants.
	\end{lemma}
	
	Since both $\hat{S}^h$ and $S$ are in $\mathbb{D}$, by the definition of $\tilde{\alpha}$ and $\check{\alpha}$, we have
	\begin{equation}
	\begin{aligned}
	(P-P_n)(\mathcal{L}'(\tilde{Y};\langle \hat{S}^h, \tilde{X}\rangle))\langle\hat{S}^h-S, \tilde{X}\rangle  \leq \tilde{\alpha}(\|\hat{S}^h-S\|_{L_2(\tilde{\Pi})}; \|\mathcal{P}^{\perp}_L\hat{S}^h\|_1;\|\mathcal{P}_L(\hat{S}^h-S)\|_1), \label{tilde}
	\end{aligned}
	\end{equation}
	and 
	\begin{equation}
	\begin{aligned}
	(P-P_n)(\mathcal{L}'(\tilde{Y};\langle \hat{S}^h, \tilde{X}\rangle))\langle\hat{S}^h-S, \tilde{X}\rangle  \leq \check{\alpha}(\|\hat{S}^h-S\|_{L_2(\tilde{\Pi})}; \|\hat{S}^h-S\|_1), \label{ProofThm_1_7}
	\end{aligned}
	\end{equation}
	If $\hat{S}^h-S \in \mathcal{K}(\mathbb{D};L;b)$, by the definition of $\alpha$, we have
	\begin{equation}
	\begin{aligned}
	(P-P_n)(\mathcal{L}'(\tilde{Y};\langle \hat{S}^h, \tilde{X}\rangle))\langle\hat{S}^h-S, \tilde{X}\rangle  \leq \alpha(\|\hat{S}^h-S\|_{L_2(\tilde{\Pi})}; \|\mathcal{P}^{\perp}_L\hat{S}^h\|_1), \label{alpha}
	\end{aligned}
	\end{equation}
	Assume for a while that
	\begin{equation}
	\|\hat{S}^h-S\|_{L_2(\tilde{\Pi})}\in [\delta_1^-, \delta_1^+],~\|\mathcal{P}_L^{\perp}\hat{S}^h\|_1\in [\delta_2^-, \delta_2^+],~\|\mathcal{P}_L^{\perp}(\hat{S}^h-S)\|_1\in [\delta_3^-, \delta_3^+]. \label{ProofThm_1_14}
	\end{equation}
	By the definition of subdifferential, for any $\hat{V}\in\partial \|\hat{S}^h\|_1$,
	$$
	\langle \hat{V}, S-\hat{S}^h\rangle \leq \|S\|_1 - \|\hat{S}^h\|_1.
	$$
	Then we apply (\ref{ProofThm_1_7}) in bound (\ref{ProofThm_1_3}) and use the upper bound on $\check{\alpha}_n(\delta_1,\delta_4)$ of Lemma \ref{Lemma_1}, and get with probability at least $1-e^{-\eta}$,
	\begin{equation}
	\begin{aligned}
	&P(\mathcal{L}(\tilde{Y};\langle \hat{S}^h, \tilde{X}\rangle)) + \|\hat{S}^h-S\|_{L_2(\tilde{\Pi})}^2 \\
	&\leq P(\mathcal{L}(\tilde{Y};\langle S, \tilde{X}\rangle))+ \varepsilon (\|S\|_1 - \|\hat{S}^h\|_1) + \check{\alpha}_n(\|\hat{S}^h-S\|_{L_2(\tilde{\Pi})},\|\hat{S}^h-S \|_1)\\
	&\leq P(\mathcal{L}(\tilde{Y};\langle S, \tilde{X}\rangle))+ \varepsilon (\|S\|_1 - \|\hat{S}^h\|_1) \\
	&+ \frac{C_3(\ell+1)R(T)\Phi a}{\sqrt{h}} \Big\{ \mathbb{E}\|\Xi\|\|\hat{S}^h-S \|_1 +\frac{2(\ell+1)R(T)\Phi a \check{\eta}}{n\sqrt{h}}+  \|\hat{S}^h-S\|_{L_2(\tilde{\Pi})}\sqrt{\frac{\check{\eta}}{n}}\Big\}.  \label{ProofThm_1_8}
	\end{aligned}
	\end{equation}
	Assuming that
	\begin{equation}
	\varepsilon >  \frac{C(\ell+1)R(T)\Phi a}{\sqrt{h}}\mathbb{E}\|\Xi\|, \label{ProofThm_1_9}
	\end{equation}
	where $C = C_1\vee 4 C_2 \vee C_3$.
	From (\ref{ProofThm_1_8})
	\begin{equation}
	P(\mathcal{L}(\tilde{Y};\langle \hat{S}^h, \tilde{X}\rangle)) \leq P(\mathcal{L}(\tilde{Y};\langle S, \tilde{X}\rangle))+ 2 \varepsilon \|S\|_1 + \frac{C_3(\ell+1)^2R(T)^2\Phi^2 a^2 \tilde{\eta}}{nh}.  \label{ProofThm_1_13}
	\end{equation}
	
	We now apply the upper bound on $\tilde{\alpha}_n(\|\hat{S}^h-S\|_{L_2(\tilde{\Pi})},\|\mathcal{P}_L^{\perp}\hat{S}^h)\|_1, \|\mathcal{P}_L(\hat{S}^h-S)\|_1) $ to (\ref{ProofThm_1_5}) and get
	with probability at least $1-e^{-\eta}$,
	\begin{equation}
	\begin{aligned}
	&P(\mathcal{L}(\tilde{Y};\langle \hat{S}^h, \tilde{X}\rangle)) + \|\hat{S}^h-S\|_{L_2(\tilde{\Pi})}^2 +  \varepsilon \|\mathcal{P}_L^{\perp}(\hat{S}^h)\|_1\\
	&\leq P(\mathcal{L}(\tilde{Y};\langle S, \tilde{X}\rangle)) + \varepsilon \|\mathcal{P}_L(\hat{S}^h-S)\|_1 + \tilde{\alpha}_n(\|\hat{S}^h-S\|_{L_2(\tilde{\Pi})},\|\mathcal{P}_L^{\perp}\hat{S}^h)\|_1, \|\mathcal{P}_L(\hat{S}^h-S)\|_1) \\
	&\leq P(\mathcal{L}(\tilde{Y};\langle S, \tilde{X}\rangle)) + \varepsilon \|\mathcal{P}_L(\hat{S}^h-S)\|_1 \\
	&+ \frac{C_2(\ell+1)R(T)\Phi a}{\sqrt{h}}\Big\{\mathbb{E}\|\Xi\|(\|\mathcal{P}_L^{\perp}\hat{S}^h)\|_1+\|\mathcal{P}_L(\hat{S}^h-S)\|_1) \Big\}+ \frac{C_2 (\ell+1)^2R(T)^2\Phi^2 a^2 \tilde{\eta}}{nh}, \label{ProofThm_1_10}
	\end{aligned}
	\end{equation}
	where the first inequality is due to the fact that 
	$$
	|\langle {\rm sign}(S), S-\hat{S}^h\rangle| =  |\langle {\rm sign}(S), \mathcal{P}_{L}(S-\hat{S}^h)\rangle| \leq \|{\rm sign}(S)\|\|\mathcal{P}_{L}(S-\hat{S}^h)\|_1 \leq \|\mathcal{P}_{L}(S-\hat{S}^h)\|_1.
	$$
	With assumption (\ref{ProofThm_1_9}) holds, we get from (\ref{ProofThm_1_10})
	\begin{equation}
	\begin{aligned}
	&P \mathcal{L}(\tilde{Y};\langle \hat{S}^h, \tilde{X}\rangle) +  \varepsilon \|\mathcal{P}_L^{\perp}(\hat{S}^h)\|_1\\
	&\leq P \mathcal{L}(\tilde{Y};\langle S, \tilde{X}\rangle) + \frac{5\varepsilon}{4} \|\mathcal{P}_L(\hat{S}^h-S)\|_1 + \frac{\varepsilon}{4} \|\mathcal{P}_L^{\perp}(\hat{S}^h)\|_1 + 
	\frac{C_2 (\ell+1)^2R(T)^2\Phi^2 a^2 \tilde{\eta}}{nh}  . \label{417}
	\end{aligned}
	\end{equation}
	If the following is satisfied:
	\begin{equation}
	\frac{C_2 (\ell+1)^2R(T)^2\Phi^2 a^2 \tilde{\eta}}{nh} \geq \frac{5\varepsilon}{4} \|\mathcal{P}_L(\hat{S}^h-S)\|_1 + \frac{\varepsilon}{4} \|\mathcal{P}_L^{\perp}(\hat{S}^h)\|_1,\label{418}
	\end{equation}
	we can just conclude that
	\begin{equation}
	P(\mathcal{L}(\tilde{Y};\langle \hat{S}^h, \tilde{X}\rangle)) \leq P(\mathcal{L}(\tilde{Y};\langle S, \tilde{X}\rangle)) + \frac{C_2 (\ell+1)^2R(T)^2\Phi^2 a^2 \tilde{\eta}}{nh}, \label{ProofThm_1_11}
	\end{equation}
	which is sufficient to meet the bound of Lemma \ref{Thm_1}. Otherwise, by the assumption that $P(\mathcal{L}(\tilde{Y};\langle \hat{S}^h, \tilde{X}\rangle)) > P(\mathcal{L}(\tilde{Y};\langle S, \tilde{X}\rangle))$, one can easily check that
	$$
	\|\mathcal{P}_L^{\perp}(\hat{S}^h-S)\|_1 \leq 5\|\mathcal{P}_L(\hat{S}^h-S)\|_1,
	$$
	which implies that $\hat{S}^h-S \in \mathcal{K}(\mathbb{D};L;5)$. This fact allows us to use the bound on $\alpha_n(\delta_1,\delta_2)$ of Lemma \ref{Lemma_1}. We get from (\ref{ProofThm_1_5})
	\begin{equation}
	\begin{aligned}
	&P(\mathcal{L}(\tilde{Y};\langle \hat{S}^h, \tilde{X}\rangle)) + \|\hat{S}^h-S\|_{L_2(\tilde{\Pi})}^2 +  \varepsilon \|\mathcal{P}_L^{\perp}(\hat{S}^h)\|_1\\
	& \leq P(\mathcal{L}(\tilde{Y};\langle S, \tilde{X}\rangle)) + \varepsilon\langle {\rm sign}(S),S- \hat{S}^h\rangle\\
	&  + \frac{C_1 (\ell+1)R(T)\Phi a }{\sqrt{h}}  \mathbb{E}\|\Xi\|(\sqrt{{\rm rank}(S)}m\| \hat{S}^h-S\|_{L_2(\tilde{\Pi})}+ \|\mathcal{P}_L^{\perp}(\hat{S}^h)\|_1) \\
	&+  \frac{C_1 (\ell+1)^2R(T)^2\Phi^2 a^2 \bar{\eta}}{nh}.
	\end{aligned}
	\end{equation}
	By applying the inequality
	$$
	\big|\langle {\rm sign}(S),\hat{S}^h-S\rangle\big|\leq m\sqrt{{\rm rank}(S)}\| \hat{S}^h-S\|_{L_2(\tilde{\Pi})},
	$$
	and the assumption (\ref{ProofThm_1_9}), we have with probability at least $1-e^{-\eta}$,
	\begin{equation}
	\begin{aligned}
	P(\mathcal{L}(\tilde{Y};\langle \hat{S}^h, \tilde{X}\rangle)) \leq  P(\mathcal{L}(\tilde{Y};\langle S, \tilde{X}\rangle))  + \varepsilon^2 m^2{\rm rank}(S) + \frac{C_1 (\ell+1)^2R(T)^2\Phi^2 a^2 \bar{\eta}}{nh}. \label{ProofThm_1_12}
	\end{aligned}
	\end{equation}
	
	To sum up, the bound of Lemma \ref{Thm_1} follows from (\ref{ProofThm_1_13}), (\ref{ProofThm_1_11}) and (\ref{ProofThm_1_12}) provided that condition (\ref{ProofThm_1_9}) and condition (\ref{ProofThm_1_14}) hold.
	
	We still need to specify $\delta_k^-$, $\delta_k^+$, $k=1,2,3,4$ to establish the bound of the theorem. By the definition of $\hat{S}^h$, we have
	$$
	P_n(\mathcal{L}(\tilde{Y};\langle X, \hat{S}^h\rangle)) + \varepsilon\| \hat{S}^h \|_{1} \leq P_n(\mathcal{L}(\tilde{Y};\langle X, 0\rangle)) \leq Q,  
	$$
	implying that $\| \hat{S}^h \|_{1} \leq \frac{Q}{\varepsilon}$. Next, 
	$
	\|\mathcal{P}_L^{\perp}\hat{S}^h \|_1 \leq \|\hat{S}^h \|_1 \leq \frac{Q}{\varepsilon}
	$
	and 
	$
	\|\mathcal{P}_L(\hat{S}^h - S) \|_1 \leq 2\|\hat{S}^h - S \|_1 \leq \frac{2Q}{\varepsilon} +2\|S\|_1.
	$
	Finally, we have $\|\hat{S}^h - S\|_{L_2(\tilde{\Pi})} \leq 2a$. Thus, we can take 
	$
	\delta_1^+:=2a$, $\delta_2^+:=\frac{Q}{\varepsilon}$, $\delta_3^+:=\frac{2Q}{\varepsilon}+2\|S\|_1$, $\delta_4^+:=\frac{Q}{\varepsilon}+\|S\|_1.
	$
	With these choices, $\delta_k^+,~k=1,2,3,4$ are upper bounds on the corresponding norms in  condition (\ref{ProofThm_1_14}). We choose 
	$\delta_1^-:=\frac{a}{\sqrt{n}},$ $\delta_2^-:=\frac{a^2}{n\varepsilon}\wedge \frac{\delta_2^+}{2}$, $\delta_3^-:=\frac{a^2}{n\varepsilon}\wedge \frac{\delta_3^+}{2}$, $\delta_4^-:=\frac{a^2}{n\varepsilon}\wedge \frac{\delta_4^+}{2}.$
	Let 
	$
	\eta^*:= \eta+3\log(B\log_2(\|S\|_1 \vee n \vee \varepsilon \vee a^{-1} \vee Q)).
	$
	It is easy to verify that $
	\bar{\eta}\vee\tilde{\eta}\vee \tilde{\eta} \leq \eta^*.
	$
	for a proper choice of numerical constant $B$ in the definition of $\eta^*$. When condition (\ref{ProofThm_1_14}) does not hold, which means at least one of the numbers $\delta_k^-,~k=1,2,3,4$ we chose is not a lower bound 
	on the corresponding norm, we can still use the bounds
	\begin{equation}
	\begin{aligned}
	&(P-P_n)(\mathcal{L}'(Y;\langle \hat{S}^h, \tilde{X}\rangle))\langle\hat{S}^h-S, \tilde{X}\rangle \\
	& \leq \tilde{\alpha}(\|\hat{S}^h-S\|_{L_2(\tilde{\Pi})}\vee \delta_1^-; \|\mathcal{P}^{\perp}_L\hat{S}^h\|_1\vee \delta_2^-;\|\mathcal{P}_L(\hat{S}^h-S)\|_1\vee \delta_3^-),
	\end{aligned}
	\end{equation}
	and 
	\begin{equation}
	\begin{aligned}
	(P-P_n)(\mathcal{L}'(Y;\langle \hat{S}^h, \tilde{X}\rangle))\langle\hat{S}^h-S, \tilde{X}\rangle \leq \check{\alpha}(\|\hat{A}(t)^{\varepsilon}-S\|_{L_2(\tilde{\Pi})}\vee \delta_1^-; \|\hat{S}^h-S\|_1\vee \delta_4^-), \label{check}
	\end{aligned}
	\end{equation}
	instead of (\ref{tilde}), (\ref{ProofThm_1_7}). In the case when $\hat{S}^h-S \in \mathcal{K}(\mathbb{D};L;5)$, we can use the bound 
	\begin{equation}
	\begin{aligned}
	(P-P_n)(\mathcal{L}'(Y;\langle \hat{S}^h, \tilde{X}\rangle))\langle\hat{S}^h-S, \tilde{X}\rangle  \leq \alpha(\|\hat{S}^h-S\|_{L_2(\tilde{\Pi})}\vee \delta_1^-; \|\mathcal{P}^{\perp}_L\hat{S}^h\|_1\vee \delta_2^-),
	\end{aligned}
	\end{equation}
	instead of bound (\ref{alpha}). Then one can repeat the arguments above with only minor modifications. By the adjusting the constants, the result of Lemma \ref{Thm_1} holds.
	
	The last thing we need to specify is the size of $\varepsilon$ which controls the nuclear norm penalty. Recall that from condition (\ref{ProofThm_1_9}), the essence is to control $\mathbb{E}\|\Xi\|$. Here we use a simple but powerful noncommutative matrix Bernstein inequalities. The original approach was introduced by \cite{ahlswede2002strong}. 
	Later, the result was improved by \cite{tropp2012user} based on the classical result of \cite{lieb1973convex}. We give the following lemma which is a direct consequence of the result proved by \cite{tropp2012user}, and we omit the proof here.
	\begin{lemma}
		\label{Lemma_2}
		Under the model (\ref{Model}), $\Xi$ is defined as in (\ref{xi}) with $\tau_j$ are
		i.i.d. uniformly distributed in [0,1], and $\varepsilon_j$ are i.i.d. Rademacher random variables, and $X_j$ are i.i.d uniformly distributed in $\mathcal{X}$. 
		Then for any $\eta>0$, with probability at least $1-e^{-\eta}$
		$$
		\|\Xi\| \leq 4 \Big(\sqrt{\frac{(\eta+\log2m)}{nm}} \bigvee \frac{(\eta+\log2m) \Phi }{n\sqrt{h}}\Big),
		$$
		and by integrating its exponential tail bounds
		$$
		\mathbb{E}\|\Xi\| \leq C \Big(\sqrt{\frac{\log2m}{nm }} \bigvee \frac{(\log2m)\Phi}{n\sqrt{h}}\Big)
		$$
		where $C$ is a numerical constant.
	\end{lemma}
	Together with (\ref{ProofThm_1_9}), we know for some numerical constant $D>0$,
	$$
	\varepsilon \geq D\frac{\Phi a(\ell+1)R(T)}{\sqrt{h}}  \Big(\sqrt{\frac{\log2m}{nm }} \bigvee \frac{(\log2m)\Phi}{n\sqrt{h}}\Big).
	$$
	which completes the proof of Lemma \ref{Thm_1}.
\end{proof}

\subsection{Proof of Lemma \ref{Lemma_1}}
\label{Proof_Lemma_1}
\begin{proof}
	We only prove the first bound in detail, and proofs of the rest two bounds are similar with only minor modifications. By Talagrand's concentration inequality \cite{talagrand1996new}, and its Bousquet's form \cite{bousquet2002bennett}, with probability at least $1-e^{-\eta}$,
	\begin{equation}
	\label{ProofLm_1}
	\alpha_n(\delta_1,\delta_2) \leq 2\mathbb{E}\alpha_n(\delta_1,\delta_2) +  \frac{24(\ell+1)^2R(T)^2\Phi^2 a^2 \eta }{nh}+ \frac{12(\ell+1)R(T)\Phi a \delta_1 \sqrt{\eta}}{\sqrt{nh}}. \\
	\end{equation}
	By standard Rademacher symmetrization inequalities, see \cite{koltchinskii2011introduction}, Sec. 2.1, we can get
	\begin{equation}
	\begin{aligned}
	\mathbb{E}\alpha_n(\delta_1,\delta_2) \leq 4\mathbb{E}\sup\Big\{ \Big|\frac{1}{n}\sum\limits_{j=1}^n\varepsilon_j (\langle A, \tilde{X}_j\rangle-\tilde{Y}_j)\langle A-S , \tilde{X}_j\rangle\Big|:A\in\mathcal{A}(\delta_1,\delta_2)\Big\}, \label{ProofLm_2}
	\end{aligned}
	\end{equation}
	where $\{\varepsilon_j\}$ are i.i.d. Rademacher random variables independent of $\{(\tau_j,X_j,\tilde{Y}_j)\}$. Then we consider the function
	$
	f(u) = (u-y+v)u,
	$
	where $|y|\leq \frac{2\Phi a}{\sqrt{h}}$ and $|v|,~|u|\leq \frac{2(\ell+1)R(T)\Phi a }{\sqrt{h}} $. Clearly, this function has a Lipschitz constant $\frac{6(\ell+1)R(T)\Phi a }{\sqrt{h}} $. Thus by comparison inequality, see \cite{koltchinskii2011introduction}, Sec. 2.2, we can get
	\begin{equation}
	\begin{aligned}
	&\mathbb{E}\sup\Big\{ \Big|n^{-1}\sum\limits_{j=1}^n\varepsilon_j(\langle A, \tilde{X}_j\rangle-\tilde{Y}_j)\langle A-S , \tilde{X}_j\rangle\Big|:A\in\mathcal{A}(\delta_1,\delta_2)\Big\}\\
	&\leq \frac{6(\ell+1)R(T)\Phi a }{\sqrt{h}}  \mathbb{E}\sup \Big\{n^{-1}\Big|\sum\limits_{j=1}^n\varepsilon_i \langle A-S ,\tilde{X}_j\rangle\Big|: A\in\mathcal{A}(\delta_1,\delta_2)\Big\}. \label{ProofLm_3}
	\end{aligned}
	\end{equation}
	As a consequence of (\ref{ProofLm_2}) and (\ref{ProofLm_3}), we have
	\begin{equation}
	\mathbb{E}\alpha_n(\delta_1,\delta_2) \leq \frac{12(\ell+1)R(T)\Phi a }{\sqrt{h}}  \mathbb{E}\sup \Big\{n^{-1}\Big|\sum\limits_{j=1}^n\varepsilon_i \langle A-S , \tilde{X}_j\rangle\Big|: A\in \mathcal{A}(\delta_1,\delta_2) \Big\}. \label{ProofLm_4}
	\end{equation}	
	The next step is to get an upper bound on
	$
	\big|n^{-1}\sum\limits_{j=1}^n\varepsilon_i \langle A-S , \tilde{X}_j\rangle\big|.
	$
	Recall that
	$
	\Xi :=n^{-1}\sum\limits_{j=1}^n\varepsilon_j  \tilde{X}_j,
	$
	then we have 
	$
	n^{-1}\sum\limits_{j=1}^n\varepsilon_i  \langle A-S , \tilde{X}_j\rangle = \langle A-S , \Xi \rangle.
	$
	One can check that
	$$
	\begin{aligned}
	|\langle A-S , \Xi \rangle| &\leq |\langle \mathcal{P}_L (A-S ), \mathcal{P}_L\Xi \rangle| + |\langle \mathcal{P}_L^{\perp} (A-S) ,\Xi \rangle|\\
	&\leq \| \mathcal{P}_L\Xi\|_2 \| \mathcal{P}_L(A-S)\|_2 + \| \Xi \| \| \mathcal{P}_L^{\perp}A\|_1\\
	&\leq m \sqrt{2{\rm rank}(S)}\| \Xi \| \|A-S\|_{L_2(\tilde{\Pi})} + \| \Xi \| \| \mathcal{P}_L^{\perp}A\|_1.
	\end{aligned}
	$$
	The second line of this inequality is due to H\"{o}lder's inequality and the third line is due to the facts that $(A-S)\in \mathcal{K}(\mathbb{D};L;5)$,
	${\rm rank}(\mathcal{P}_L(\Xi))\leq 2{\rm rank}(S)$, $\| \mathcal{P}_L\Xi\|_2 \leq 2\sqrt{{\rm rank}(\mathcal{P}_L(\Xi))}\|\Xi\|,$
	and $\big \|A-S \big\|_{L_2(\tilde{\Pi})}^2 =  \frac{1}{m^2}\big\| A-S \big\|_2.$
	Therefore,
	\begin{equation}
	\begin{aligned}
	&  \frac{12(\ell+1)R(T)\Phi a }{\sqrt{h}} \mathbb{E}\sup \Big\{\Big|n^{-1}\sum\limits_{j=1}^n\varepsilon_i \langle A-S , \tilde{X}_j\rangle\Big|: A\in \mathcal{A}(\delta_1,\delta_2) \Big\} \\
	&\leq  \frac{12(\ell+1)R(T)\Phi a }{\sqrt{h}} \mathbb{E}\|\Xi\|(2\sqrt{2{\rm rank}(S)} m\delta_1+\delta_2). \label{ProofLm_5}
	\end{aligned}
	\end{equation}
	It follows from (\ref{ProofLm_1}), (\ref{ProofLm_4}) and (\ref{ProofLm_5}) that with probability at least $1-e^{-\eta}$,
	\begin{align*}
	\alpha_n(\delta_1,\delta_2)  &\leq  \Big(  \frac{12(\ell+1)R(T)\Phi a }{\sqrt{h}} \mathbb{E}\|\Xi\|(\sqrt{{\rm rank}(S)} m\delta_1+\delta_2)\Big) \\
	&+  \frac{24(\ell+1)^2R(T)^2\Phi^2 a^2 \eta }{nh}+ \frac{12(\ell+1)R(T)\Phi a \delta_1 \sqrt{\eta}}{\sqrt{nh}}. 
	\end{align*}
	Now similar to the approach in \cite{koltchinskii2013sharp}, we make this bound uniform in $\delta_k\in [\delta_k^-, \delta_k^+].$
	Let 
	$
	\delta_k^{j_k} = \delta_k^+2^{-j_k},j_k=0,...,[\log_2(\delta^+_k/\delta_k^-)]+1,~k=1,2.
	$
	By the union bound, with probability at least $1-e^{-\eta}/3$, for all $j_k=0,...,[\log_2(\delta^+_k/\delta_k^-)]+1,~k=1,2$, we have
	\begin{align*}
	\alpha_n(\delta_1,\delta_2) & \leq  \Big(  \frac{12(\ell+1)R(T)\Phi a }{\sqrt{h}} \mathbb{E}\|\Xi\|(\sqrt{{\rm rank}(S)} m\delta_1^{j_1}+\delta_2^{j_2})\Big)\\
	& +  \frac{24(\ell+1)^2R(T)^2\Phi^2 a^2 \eta }{nh}+ \frac{12(\ell+1)R(T)\Phi a \delta_1^{j_1} \sqrt{\eta}}{\sqrt{nh}}. 
	\end{align*}
	which implies that for all $\delta_k\in[\delta_k^-,\delta_k^+],~k=1,2$,
	\begin{align*}
	\alpha_n(\delta_1,\delta_2) & \leq  \Big(  \frac{12(\ell+1)R(T)\Phi a }{\sqrt{h}} \mathbb{E}\|\Xi\|(\sqrt{{\rm rank}(S)} m\delta_1+\delta_2)\Big)\\
	& +  \frac{24(\ell+1)^2R(T)^2\Phi^2 a^2 \bar{\eta} }{nh}+ \frac{12(\ell+1)R(T)\Phi a \delta_1 \sqrt{\bar{\eta}}}{\sqrt{nh}}. 
	\end{align*}
	The proofs of the second and the third bounds are similar to this one, we omit the repeated arguments.
\end{proof}

\end{document}